%% file: main.tex
\documentclass{article}
\usepackage[utf8]{inputenc}
\usepackage{authblk}
\usepackage{amsfonts} 
\usepackage{amsmath}
\usepackage[maxbibnames=99]{biblatex}
\usepackage{color}
\usepackage[colorlinks]{hyperref}
\usepackage{subcaption}
\usepackage{graphicx}

\title{Procedural Humans for Computer Vision}

\renewcommand{\today}{\ifcase \month \or January\or February\or March\or %
April\or May \or June\or July\or August\or September\or October\or November\or %
December\fi~\number \year}

\ifx\anon\undefined
\author{Charlie Hewitt}
\author{Tadas Baltru\v{s}aitis}
\author{Erroll Wood}
\author{Lohit Petikam}
\author{Louis Florentin}
\author{Hanz Cuevas Velasquez}

\affil{Mesh Labs Cambridge, Microsoft}
\else
\author{Anonymized for CVPR 2023 review}
\fi

\date{\today}

\usepackage{bm}
\newcommand{\myvec}[1]{\boldsymbol{\mathbf{#1}}}

\usepackage{ulem}

\begin{document}

\maketitle

Recent work has shown the benefits of synthetic data for use in computer vision, with applications ranging from autonomous driving~\cite{sun2022shift, thomas_pandikow_kim_stanley_grieve_2021} to face landmark detection~\cite{wood2021fake} and reconstruction~\cite{ wood2022dense}.
There are a number of benefits of using synthetic data from privacy preservation and bias elimination~\cite{bae2023digiface1m, mcduff2021synthetic} to quality and feasibility of annotation~\cite{wood2022dense}.
Generating human-centered synthetic data is a particular challenge in terms of realism and domain-gap, though recent work has shown that effective machine learning models can be trained using synthetic face data alone~\cite{wood2021fake}.
We show that this can be extended to include the full body by building on the pipeline of \textcite{wood2021fake} to generate synthetic images of humans in their entirety, with ground-truth annotations for computer vision applications.

In this report we describe how we construct a parametric model of the face and body, including articulated hands; our rendering pipeline to generate realistic images of humans based on this body model; an approach for training DNNs to regress a dense set of landmarks covering the entire body; and a method for fitting our body model to dense landmarks predicted from multiple views.

\section{Shape Model}
\label{sec:shape_model}

\input{shape_model}

\section{Rendering Pipeline}
\label{sec:render_pipeline}

\input{render_pipeline}

\section{Landmark Regression}
\label{sec:landmarks}

\input{landmarks}

\section{Model fitting}
\label{sec:model_fitting}

\input{model_fitting}

\ifx\anon\undefined
\section*{Acknowledgements}

Thanks to Rodney Brunet and his team for invaluable help with clothing assets and artistic input throughout,
Darren Cosker for assistance in collecting motion-capture data for the pose library and Kendall Robertson for assistance with processing parts of the pose library.
\else
\fi

\printbibliography

\end{document}

%% file: shape_model.tex
\subsection{Model Construction}

\begin{figure}
    \centering
    \begin{subfigure}{0.32\textwidth}
        \includegraphics[width=\textwidth]{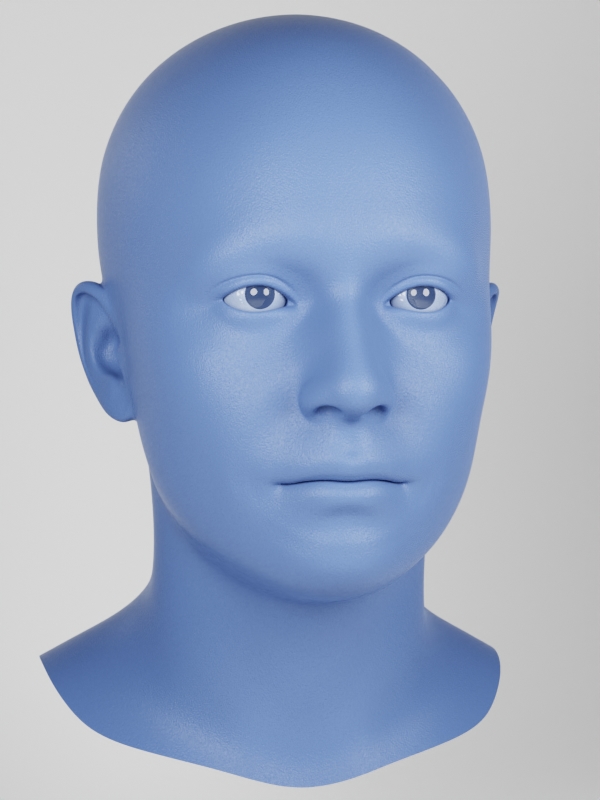}
        \caption{\centering Head model of \textcite{wood2021fake}}
    \end{subfigure}
    \begin{subfigure}{0.32\textwidth}
        \includegraphics[width=\textwidth]{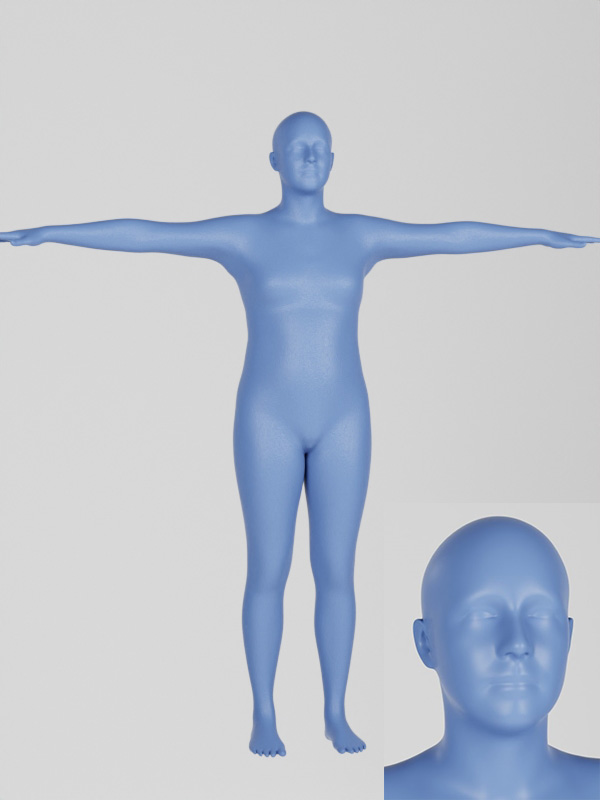}
        \caption{\centering SMPL-H body model~\cite{MANO:SIGGRAPHASIA:2017}}
    \end{subfigure}
    \begin{subfigure}{0.32\textwidth}
        \includegraphics[width=\textwidth]{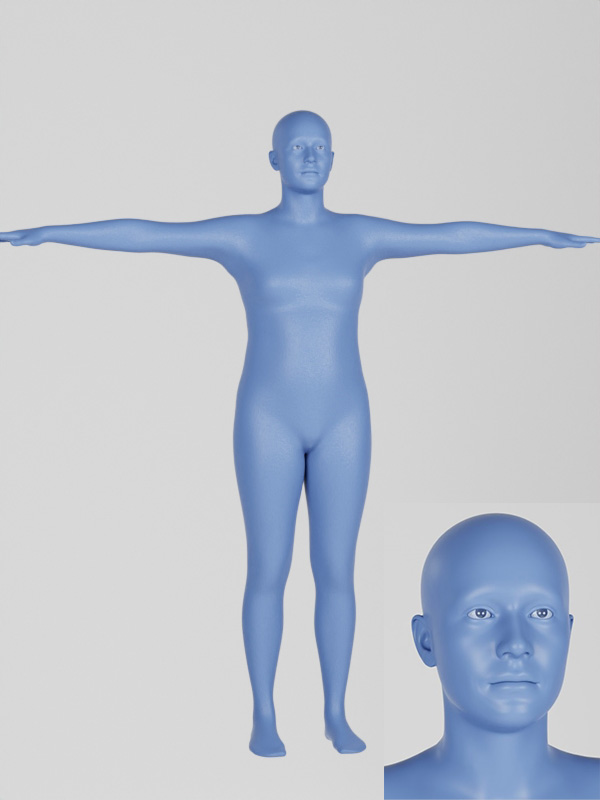}
        \caption{\centering Our combined body and head model}
    \end{subfigure}
    \caption{Template meshes of the constituent models (left two) used in our combined shape model (right), with head region inset for full-body models. Our combined model has significantly higher fidelity on the face than the SMPL-H body model~\cite{MANO:SIGGRAPHASIA:2017}.}
    \label{fig:models}
\end{figure}

Our body model combines the high fidelity face model of \textcite{wood2021fake} with the popular body and hand model SMPL-H~\cite{MANO:SIGGRAPHASIA:2017}, which itself combines the articulated hands of MANO~\cite{MANO:SIGGRAPHASIA:2017} with the SMPL body model~\cite{SMPL:2015}. So, we obtain a parametric model of the full human body with control of body shape and pose as in SMPL-H~\cite{MANO:SIGGRAPHASIA:2017}, and of facial and identity and expression as in \textcite{wood2021fake}, see \autoref{fig:models}.

The body mesh is made up of $ N = 12943 $ vertices and 12726 polygons
with a skeleton of $ K = 54 $ joints:
22 for the body (the SMPL skeleton~\cite{SMPL:2015}),
15 per hand (as in MANO/SMPL-H~\cite{MANO:SIGGRAPHASIA:2017})
and 2 for the eyes.

The body mesh vertex positions are defined by mesh generating function
$ \mathcal{M}(\vec{\gamma}, \vec{\beta}, \vec{\psi}, \vec{\theta})\!:\!\mathbb{R}^{|\vec{\gamma}|+|\vec{\beta}|+|\vec{\psi}|+|\vec{\theta}|}\!\to\!\mathbb{R}^{N\times3} $ which takes parameters
$ \vec{\gamma}\in\mathbb{R}^{|\vec{\gamma}|} $ for face identity,
$ \vec{\beta}\in\mathbb{R}^{|\vec{\beta}|} $ for body identity,
$ \vec{\psi}\in\mathbb{R}^{|\vec{\psi}|} $ for expression, and
$ \vec{\theta}\in\mathbb{R}^{K\times3} $ for skeletal pose.

\begin{equation*}
    \mathcal{M}(\vec{\gamma}, \vec{\beta}, \vec{\psi}, \vec{\theta}) =
    \mathcal{L}(\mathcal{T}(\vec{\gamma}, \vec{\beta}, \vec{\psi}, \vec{\theta}), \vec{\theta}, \mathcal{J}(\vec{\gamma}, \vec{\beta}); \mathbf{W})
\end{equation*}

where $ \mathcal{L}(\mathbf{X}, \vec{\theta}, \mathbf{J}; \mathbf{W}) $ is a standard linear blend skinning (LBS) function that rotates
vertex positions $ \mathbf{X}\in\mathbb{R}^{N\times3} $ about joint locations $ \mathbf{J}\in\mathbb{R}^{K\times3} $ by local
joint rotations $ \vec{\theta} $, with per-vertex hand-authored skinning weights $ \mathbf{W}\in\mathbb{R}^{K\times N} $ determining how rotations are interpolated across the mesh.

$ \mathcal{T}(\vec{\gamma}, \vec{\beta}, \vec{\psi}, \vec{\theta})\!:\!\mathbb{R}^{|\vec{\beta}| + |\vec{\gamma}| + |\vec{\psi}| + |\vec{\theta}|}\to\mathbb{R}^{N\times3} $
constructs an unposed body mesh by adding displacements to the template mesh $ \mathbf{\overline{T}}\!\in\!\mathbb{R}^{N \times 3} $, which represents the average body in T-pose with neutral expression:

\begin{equation*}
      \mathcal{T}(\vec{\gamma}, \vec{\beta}, \vec{\psi}, \vec{\theta})^j_{\:k} =
      \overline{T}^j_{\:k} +
      \gamma_i S^{ij}_{\;\;k} +
      \beta_i U^{ij}_{\;\;k} +
      \psi_i E^{ij}_{\;\;k} +
      P(\vec{\theta})^j_{\:k}
\end{equation*}

given linear face identity basis $ \mathbf{S}\!\in\!\mathbb{R}^{|\vec{\gamma}| \times N \times 3} $,
body identity basis $ \mathbf{U}\!\in\!\mathbb{R}^{|\vec{\beta}| \times N \times 3} $,
expression basis $ \mathbf{E}\!\in\!\mathbb{R}^{|\vec{\psi}| \times N \times 3} $ and
$ P(\vec{\theta}) $ which represents pose-dependent blendshape offsets for pose parameters $ \vec{\theta} $ (see SMPL~\cite{SMPL:2015} for more details).
Note the use of Einstein summation notation in this definition and below.
Finally, $ \mathcal{J}(\vec{\gamma}, \vec{\beta})\!:\!\mathbb{R}^{|\vec{\gamma}| + |\vec{\beta}|}\to\mathbb{R}^{K\times3} $ moves the joint locations to account for changes in identity:

\begin{equation*}
      \mathcal{J}(\vec{\gamma}, \vec{\beta})^j_{\:k} =
      J(\overline{T}^j_{\:k} + \gamma_i S^{ij}_{\;\;k} + \beta_i U^{ij}_{\;\;k})
\end{equation*}

Where $ J $ is a modified version of the SMPL joint regressor, learnt as part of the SMPL model.

The face identity, $ \mathbf{S} $, and expression $ \mathbf{E} $ bases are those from \textcite{wood2021fake},
and the body identity basis $ \mathbf{U} $ is that from the neutral SMPL-H model, which is a PCA basis learnt from scans of humans.
The pose-dependent blendshapes are also taken from the neutral SMPL-H model.

The size of the face identity basis is $ |\vec{\gamma}| = 260$ ,
and the body identity basis is $ |\vec{\beta}| = 300$.
The expression basis has $ |\vec{\psi}| = 224 $ components.

\subsubsection{Template Mesh}

\begin{figure}
    \begin{minipage}{0.5\linewidth}
        \includegraphics[width=\linewidth]{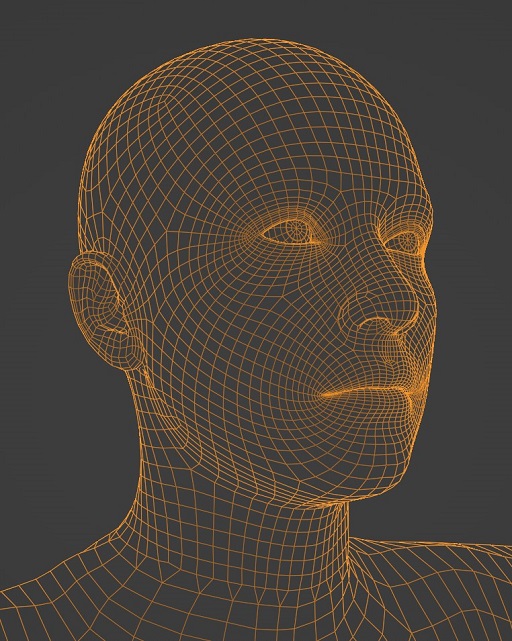}
        \caption{Head and neck topology of our body model.}
        \label{fig:neck_topo}
    \end{minipage}
    \hfill
    \begin{minipage}{0.454\linewidth}
       \includegraphics[width=\linewidth]{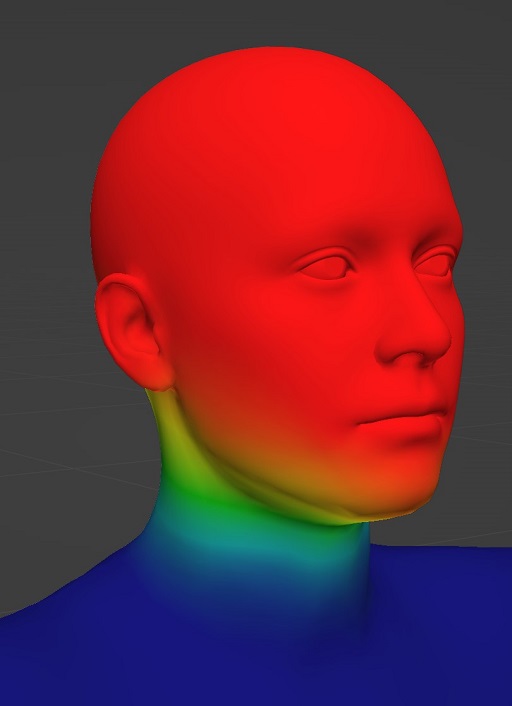}
        \caption{Head mask used for blending bases.}
        \label{fig:head_mask}
    \end{minipage}
\end{figure}

To construct the template mesh, $\overline{T}$, we manually align the template of \textcite{wood2021fake} to the head of the SMPL template.
Once aligned the head of SMPL and lower neck of the new head are removed and the two partial meshes merged.
The topology around the join was hand-crafted to create a smooth transition given the different density of the two meshes, see \autoref{fig:neck_topo}.

\subsubsection{UV Space}

To create the UV space we started from the SMPL UV space and manually aligned the UVs for the new head vertices with the original boundaries and features of the head in the SMPL UV space.
This means that the UV space of SMPL and our model are functionally identical so textures can be reused directly.
Elements that are not present in SMPL such as the eyes and mouth parts were added in previously unused areas of the UV space.


\subsubsection{Basis Transfer}

Given that we have changed the topology of the mesh significantly from both \textcite{wood2021fake} and SMPL, we need to adapt all of the bases associated with these source models to work for ours.
That is face and body identity, expression, pose-dependent blendshapes, skinning weights and the joint regressor.

We calculate a mapping function $ \mathcal{F}_{Q}\!:\!\mathbb{R}^{|Q| \times m}\to\mathbb{R}^{N \times m} $
which transforms vertex data from a given topology $Q$ to that of our model, where $ |Q| $ is the number of vertices in model $ Q $.
Specifically, we calculate $ \mathcal{F}_{head} $ and $ \mathcal{F}_{smpl} $.

This mapping function $\mathcal{F}_{Q}$ is determined by finding, for every vertex in our template mesh $\mathbf{\overline{T}}$,
the closest point on the surface of the template mesh of $Q$.
This is then stored as barycentric coordinates on the triangle which that point lies within, we can then transform the input data
for a given vertex in our model by taking the sum, weighted by barycentric coordinates, of the data from the vertices of that triangle in $Q$.
This approach works because all data (identity basis, pose basis etc.) for all models is stored per vertex.

We can then apply this mapping function to bases directly to map them from the original head or body models to our model,
e.g., the face identity basis: $ \mathbf{S} = \mathcal{F}_{head}(\mathbf{S}_{head}) $.
Vertex groups can be mapped by creating a mask containing one where a vertex in the group and zero where it isn't, and applying $\mathcal{F}$ to this mask. 
Vertices are then determined to be in the vertex group for our model if the mapped value for the vertex is above some threshold value.

In order to prevent the head identity basis affecting the lower neck area we mask it to only apply to the head.
Similarly, to prevent the SMPL identity basis and pose dependent blendshapes affecting the head, we mask these to only apply to the body.
In order to preserve variation in head position due to body identity, we take the average of the SMPL identity basis over the masked area and apply it uniformly to all masked vertices.
The mask is constructed by taking the SMPL skinning weights for the head joint, mapped to our model, and adding the eyes and mouth parts from \textcite{wood2021fake}, see \autoref{fig:head_mask}.

It is also not sufficient to simply map the SMPL joint regressor to our model using $\mathcal{F}$, as the per-joint regressor must sum to exactly one.
Instead we calculate a one-to-one vertex mapping from SMPL to our model by taking the closest vertex pairing on the two template meshes.
Given that our model is based on SMPL we get an exact match for all but the head and neck joints, where we get a very close approximation.
As we have added additional joints for the eyes, we also need to construct eye joint regressors.
We do this by simply taking the four extreme points of each eyeball in the $x$ and $y$ directions.

Similarly the skinning weights for the eyeballs need to be overridden, removing any influence from other joints and setting
the influence of the applicable eye joint for all eyeball vertices to one.
The mouth parts skinning weights are also overridden to completely follow the head joint, as the mapping function above
can result in the neck joint having some influence.

\subsection{Identity Sampling}

To sample face identity we fit a multi-variate Gaussian to male, female and all identities in the training set of \textcite{wood2021fake} (all gender labels are self-reported).
This lets us randomly sample male, female and neutral (non-binary) facial identities.

For body identity, SMPL has three versions of the model: male, female and neutral but for our model we use just one, neutral, model.
As such it is useful to be able to transfer shape parameters from gendered models to neutral, that is for gendered parameters
$ \vec{\beta}_g $, find neutral parameters $ \vec{\beta}_n $ such that $ \textrm{SMPL}_g(\vec{\beta}_g) = \textrm{SMPL}_n(\vec{\beta}_n) $
where $ \textrm{SMPL}_g $ is the mesh generating function for the SMPl model of gender $ g $.

So for template vertices $ \mathbf{T} $, shape basis $ \mathbf{S} $ and shape parameters $ \vec{\beta} $,
we want to find template offset $ \vec{o} $ and identity mapping $ \mathbf{M} $ such that

\begin{equation*}
    \mathbf{T}_g + \vec{\beta}_g \cdot \mathbf{S}_g = \mathbf{T}_n + (\vec{o}_g + \vec{\beta}_g \cdot \mathbf{M}_g) \cdot \mathbf{S}_n
\end{equation*}

To do this we solve the two sub-problems finding the least squares solution for:

\begin{align*}
    \mathbf{T}_g &= \mathbf{T}_n + \vec{o}_g \cdot \mathbf{S}_n\\
    \mathbf{T}_g - \mathbf{T}_n &= \vec{o}_g \cdot \mathbf{S}_n\\
    \vec{o} &= \mathbf{S}_n / (\mathbf{T}_g - \mathbf{T}_n)
\end{align*}

and

\begin{align*}
    \vec{\beta}_g \cdot \mathbf{S}_g &= \vec{\beta}_g \cdot \mathbf{M}_g \cdot \mathbf{S}_n\\
    \mathbf{S}_g &= \mathbf{M}_g \cdot \mathbf{S}_n\\
    M_g^i &= \mathbf{S}_n / S_g^i
\end{align*}

for the latter we solve for each element of the shape basis, $S_g^i$ individually.

So given this mapping information, $\vec{o}_g$ and $\mathbf{M}_g$, for both male and female SMPL models ($g \in [m, f]$) we can simply sample the unit normal to get identity $\vec{\beta}_g$ and for given gender, $g$, transfer to the neutral identity (usable by our model) $\vec{\beta}_n = \vec{o}_g + \vec{\beta}_g \cdot \mathbf{M}_g$.

\begin{figure}
    \centering
    \begin{subfigure}{\textwidth}
        \includegraphics[width=\textwidth]{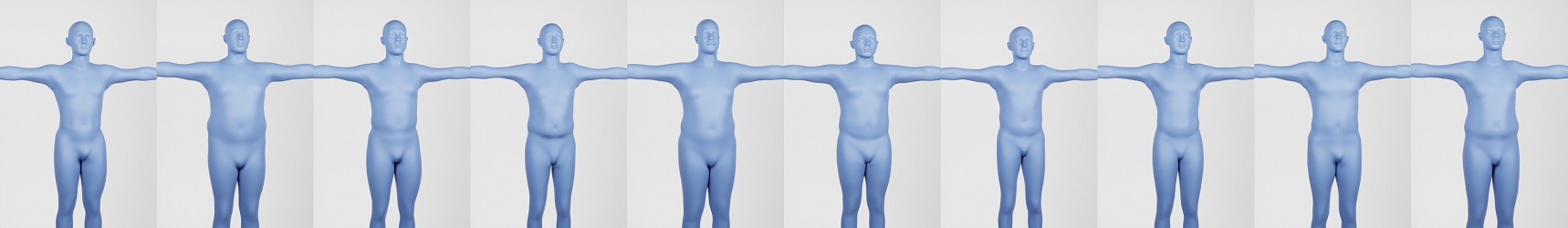}
        \caption{Male.}
    \end{subfigure}
    \begin{subfigure}{\textwidth}
        \includegraphics[width=\textwidth]{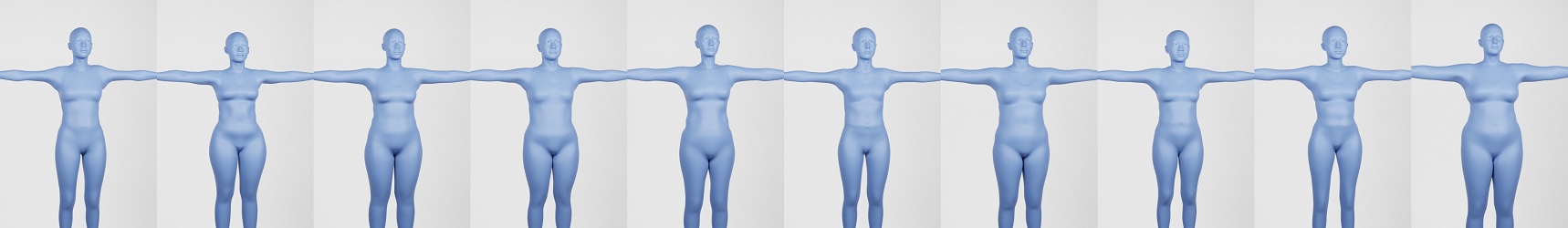}
        \caption{Female.}
    \end{subfigure}
    \begin{subfigure}{\textwidth}
        \includegraphics[width=\textwidth]{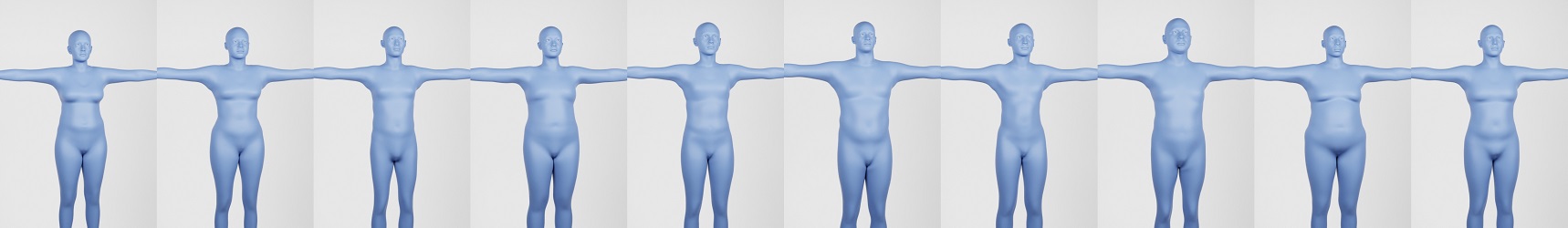}
        \caption{Neutral.}
    \end{subfigure}
    \caption{Randomly sampled identities.}
    \label{fig:identity_samples}
\end{figure}

We currently have no concept of dependence between the body and face identities, meaning there can be significant mismatch in the shape. 
In practice we find that sampling with coherent gender produces plausible results in most cases.
Joint sampling of face and body identity could be an interesting direction for future work.
Example sampled identities can be seen in \autoref{fig:identity_samples}.
In general we sample male, female and neutral identities in equal proportion to ensure we cover the gender spectrum sufficiently.

%% file: render_pipeline.tex
\begin{figure}
    \centering
    \includegraphics[width=\linewidth]{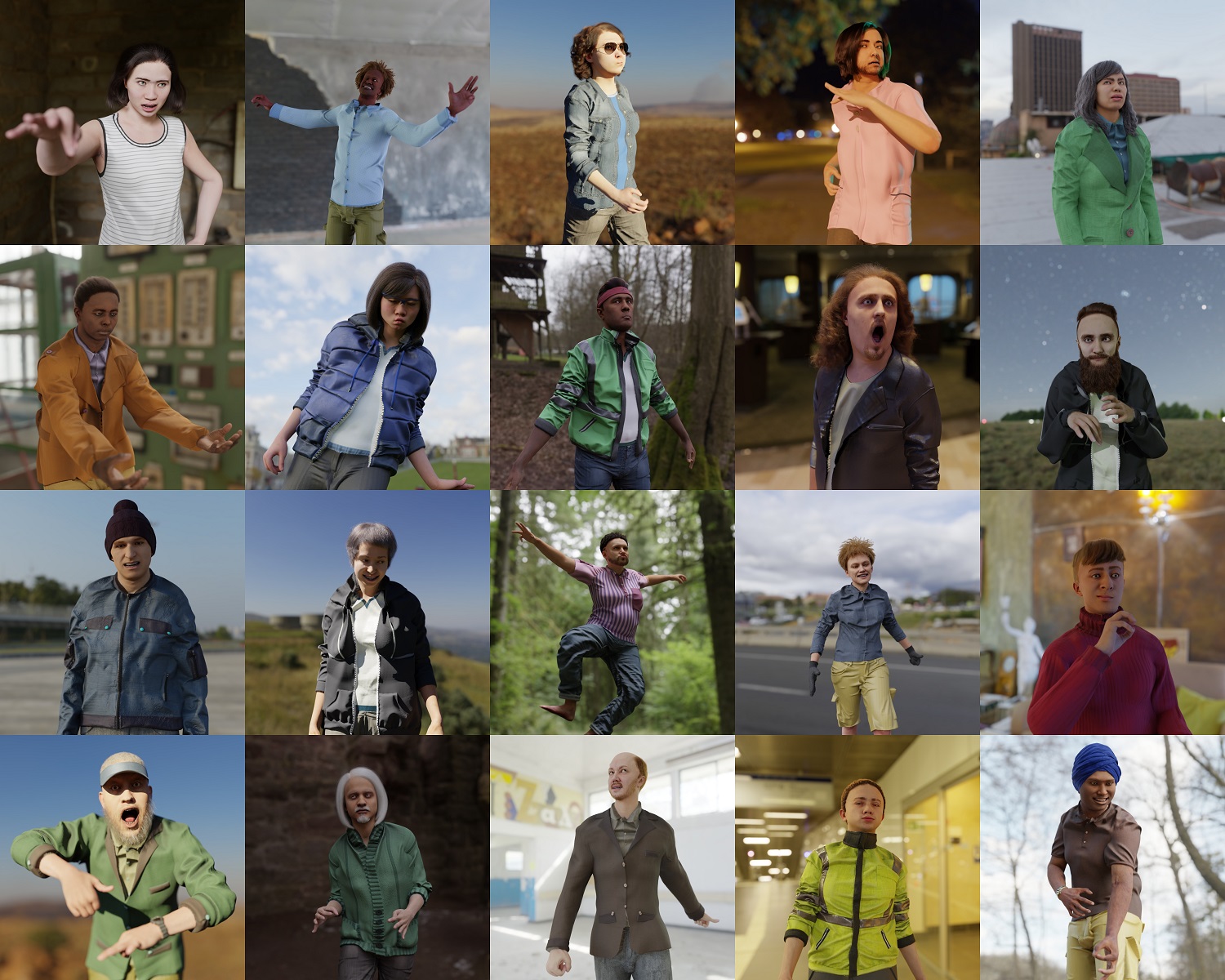}
    \caption{Example images generated using our human synthetics pipeline.}
    \label{fig:pretty_renders}
\end{figure}

We build on the pipeline of \textcite{wood2021fake} using the Cycles rendering engine~\cite{CyclesRenderer}.
Many elements are identical such as the hair and environment libraries.
We additionally add a shadow-catching plane to integrate the subject better with the scene now that the legs/feet are included.
\autoref{fig:pretty_renders} shows some of the final renders from the pipeline.

\begin{figure}
    \centering
    \begin{subfigure}{0.32\linewidth}
        \includegraphics[width=\linewidth]{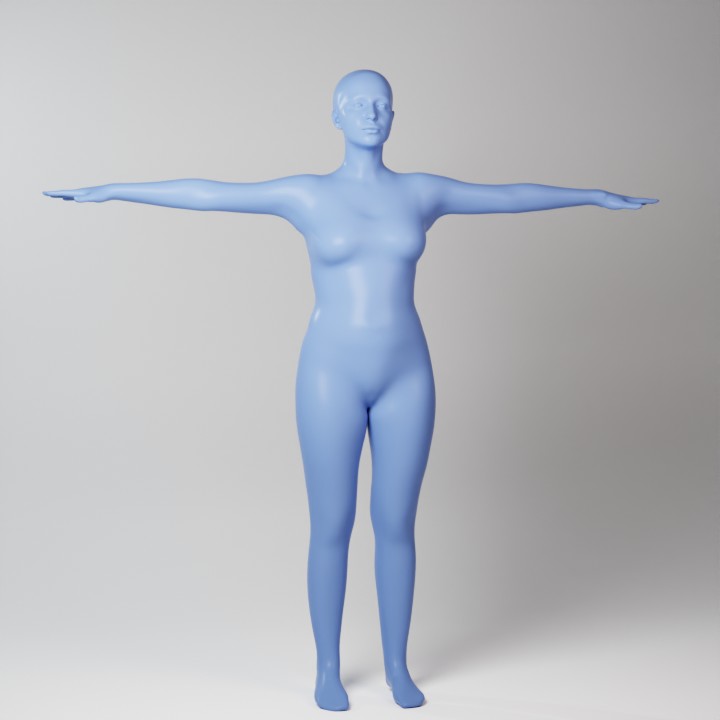}
        \caption{\centering Sampled identity.}
    \end{subfigure}
    \begin{subfigure}{0.32\linewidth}
        \includegraphics[width=\linewidth]{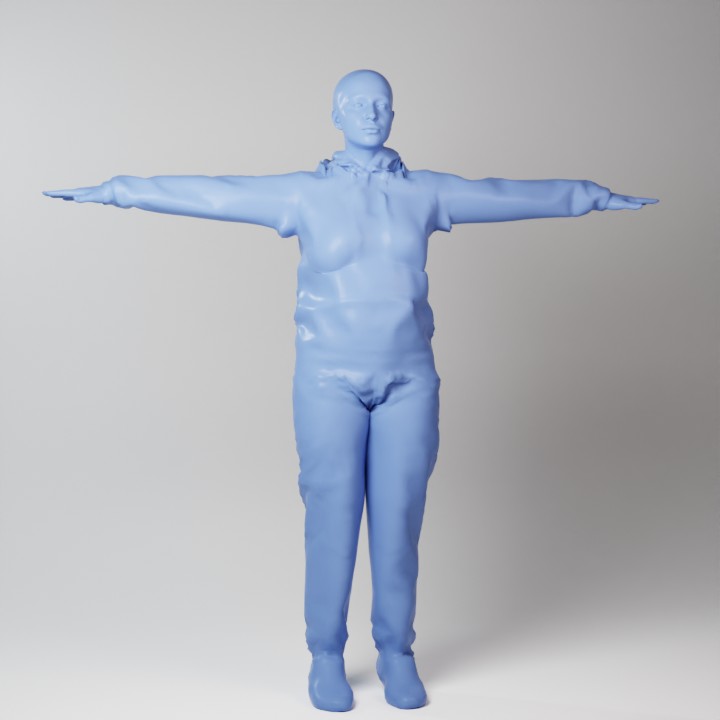}
        \caption{\centering Clothing.}
    \end{subfigure}
    \begin{subfigure}{0.32\linewidth}
        \includegraphics[width=\linewidth]{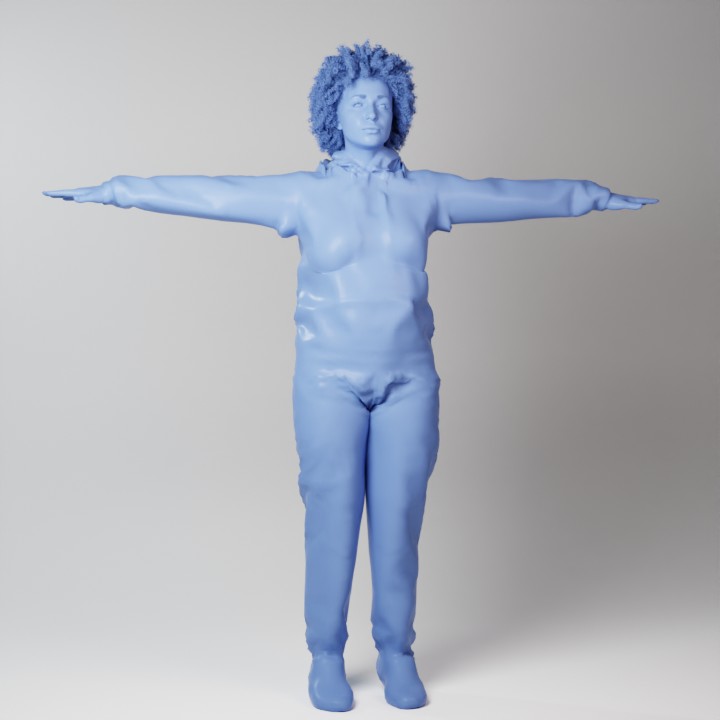}
        \caption{\centering Hair.}
    \end{subfigure}
    \begin{subfigure}{0.32\linewidth}
        \includegraphics[width=\linewidth]{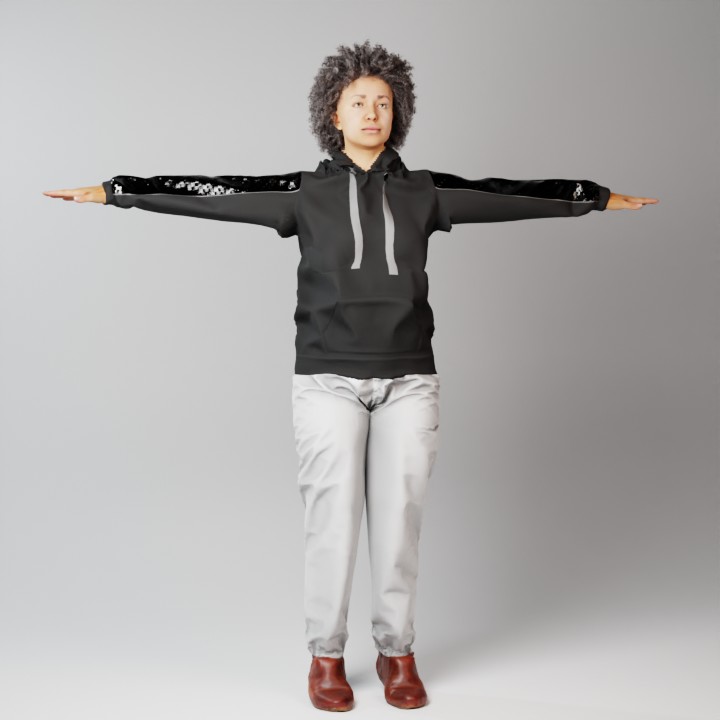}
        \caption{\centering Textures/Shaders.}
    \end{subfigure}
    \begin{subfigure}{0.32\linewidth}
        \includegraphics[width=\linewidth]{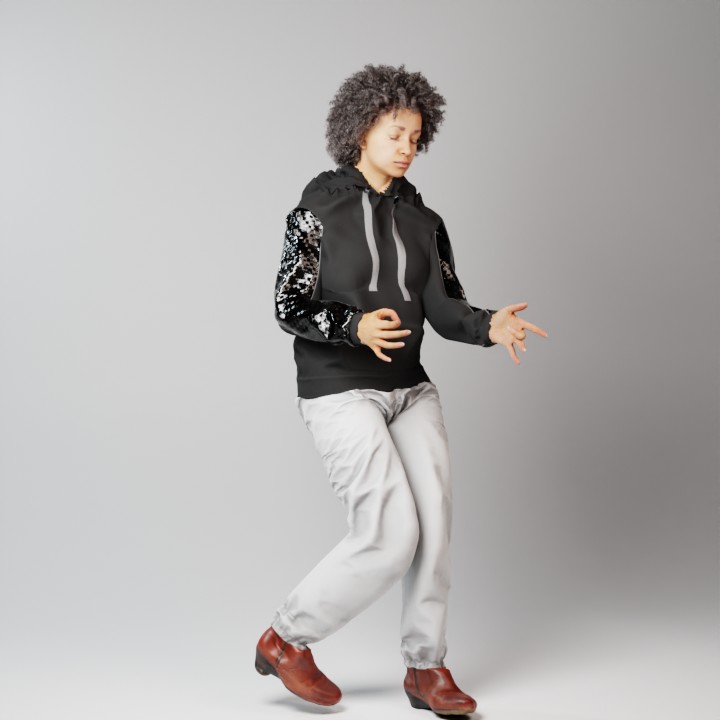}
        \caption{\centering Pose.}
    \end{subfigure}
    \begin{subfigure}{0.32\linewidth}
        \includegraphics[width=\linewidth]{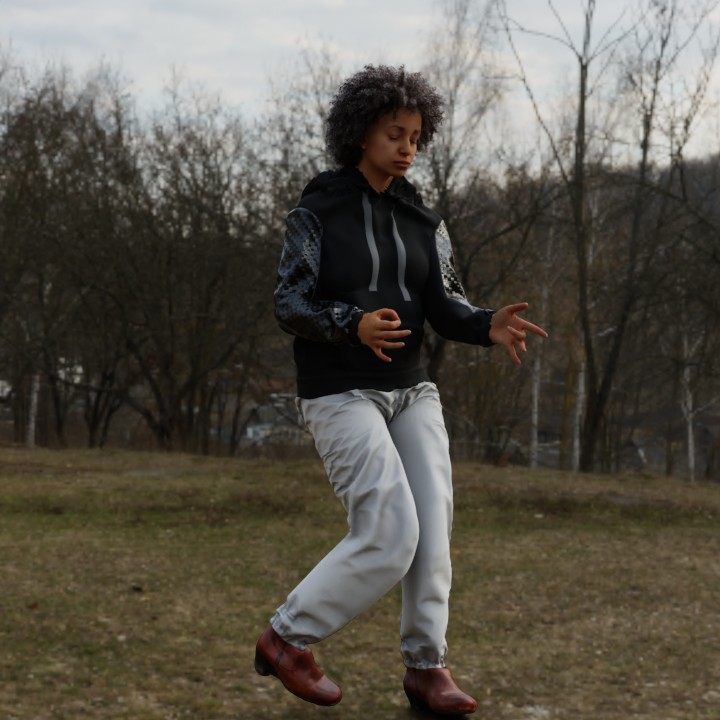}
        \caption{\centering Environment.}
    \end{subfigure}
    \caption{Stages of our pipeline to construct a synthetic human.}
    \label{fig:construction}
\end{figure}

Primary differences are the texture library used for the rest of the body, clothing for the body (which now needs to adapt to pose changes rather than being static as in \textcite{wood2021fake}), and the pose library.
Details of these additions are given in the following section.
\autoref{fig:construction} shows how we construct a synthetic image of a human from the component parts.

\begin{figure}
    \centering
    \begin{subfigure}{0.24\linewidth}
        \includegraphics[width=\linewidth]{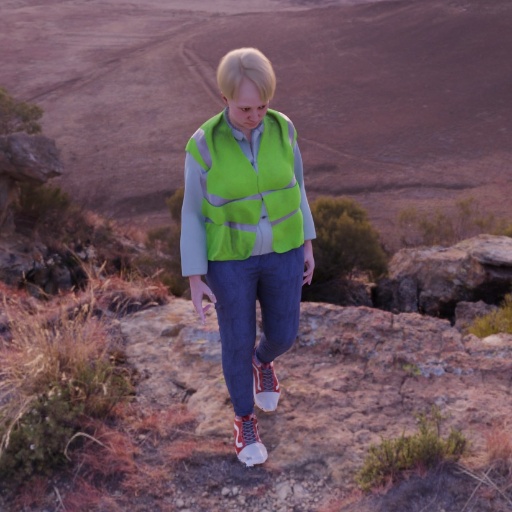}
        \caption{\centering RGB}
        \label{fig:ground_truth:rgb}
    \end{subfigure}
    \begin{subfigure}{0.24\linewidth}
        \includegraphics[width=\linewidth]{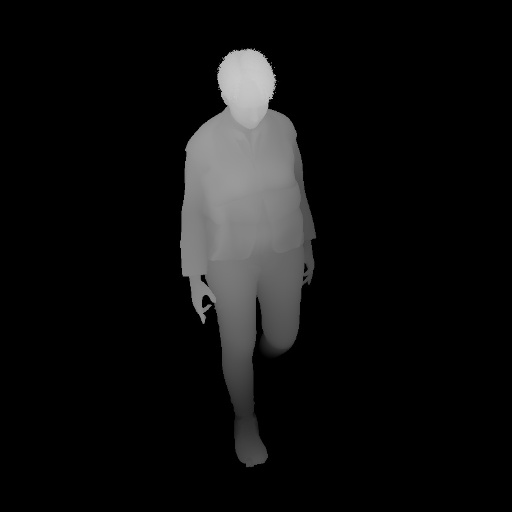}
        \caption{\centering Depth}
    \end{subfigure}
    \begin{subfigure}{0.24\linewidth}
        \includegraphics[width=\linewidth]{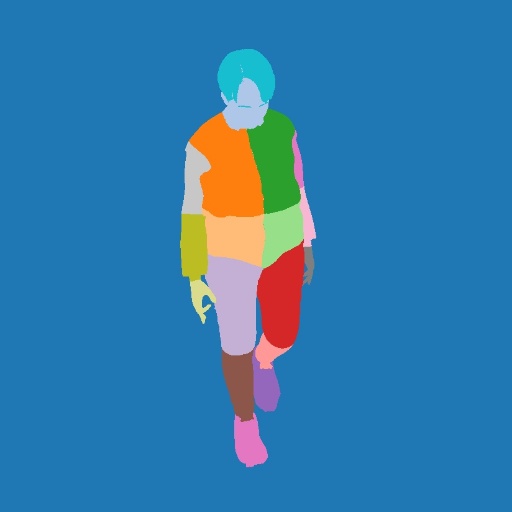}
        \caption{\centering Segmentation}
    \end{subfigure}
    \begin{subfigure}{0.24\linewidth}
        \includegraphics[width=\linewidth]{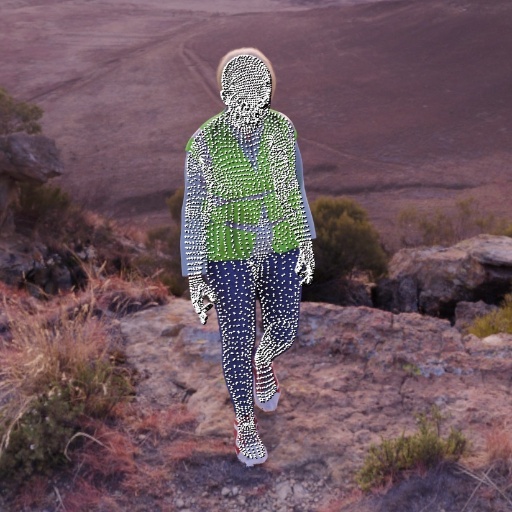}
        \caption{\centering Vertices}
        \label{fig:ground_truth:vertices}
    \end{subfigure}
    \begin{subfigure}{0.24\linewidth}
        \includegraphics[width=\linewidth]{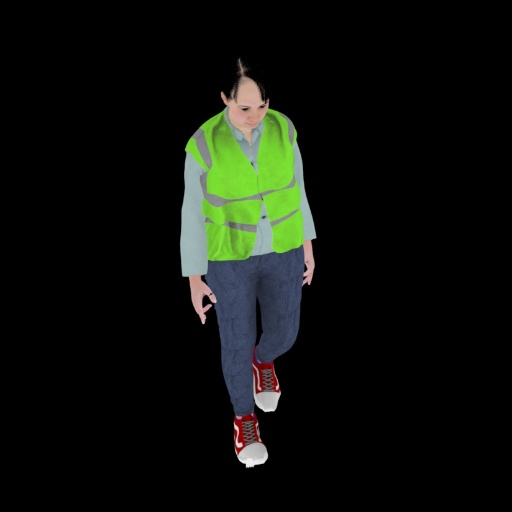}
        \caption{\centering Albedo}
    \end{subfigure}
    \begin{subfigure}{0.24\linewidth}
        \includegraphics[width=\linewidth]{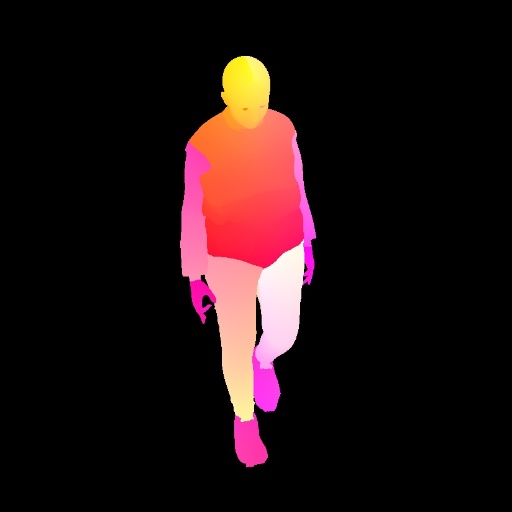}
        \caption{\centering UVs}
    \end{subfigure}
    \begin{subfigure}{0.24\linewidth}
        \includegraphics[width=\linewidth]{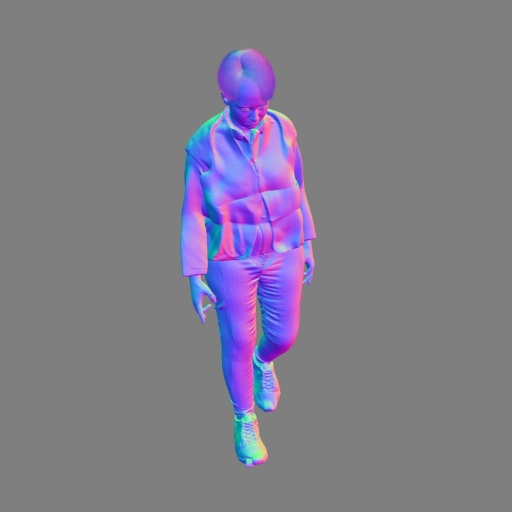}
        \caption{\centering Normals}
    \end{subfigure}
    \begin{subfigure}{0.24\linewidth}
        \includegraphics[width=\linewidth]{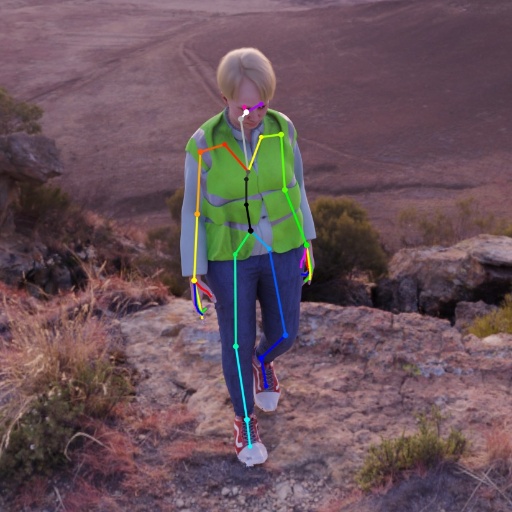}
        \caption{\centering Skeleton}
    \end{subfigure}
    \caption{Many ground truth label types can be generated using our pipeline.}
    \label{fig:ground_truth}
\end{figure}

We are able to generate a wide variety of ground truth data from our pipeline, along with RGB images, in the same fashion as  \textcite{wood2021fake}.
\autoref{fig:ground_truth} shows some example ground truth annotations for an image generated using our pipeline.

\subsection{Textures}

\begin{figure}
    \centering
        \begin{subfigure}{0.45\linewidth}
        \includegraphics[width=\linewidth]{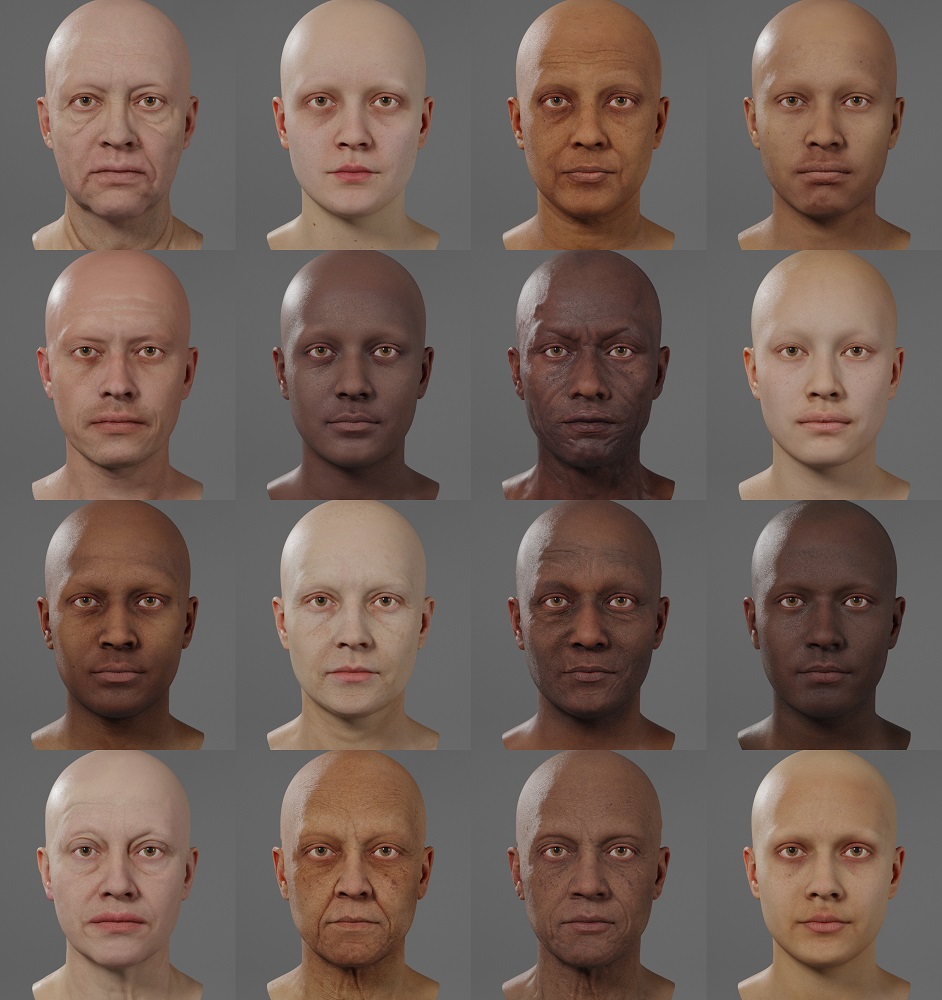}
        \caption{\centering Face~\cite{wood2021fake}.}
        \label{fig:face_skin_textures}
    \end{subfigure}
    \begin{subfigure}{0.478\linewidth}
        \includegraphics[width=\linewidth]{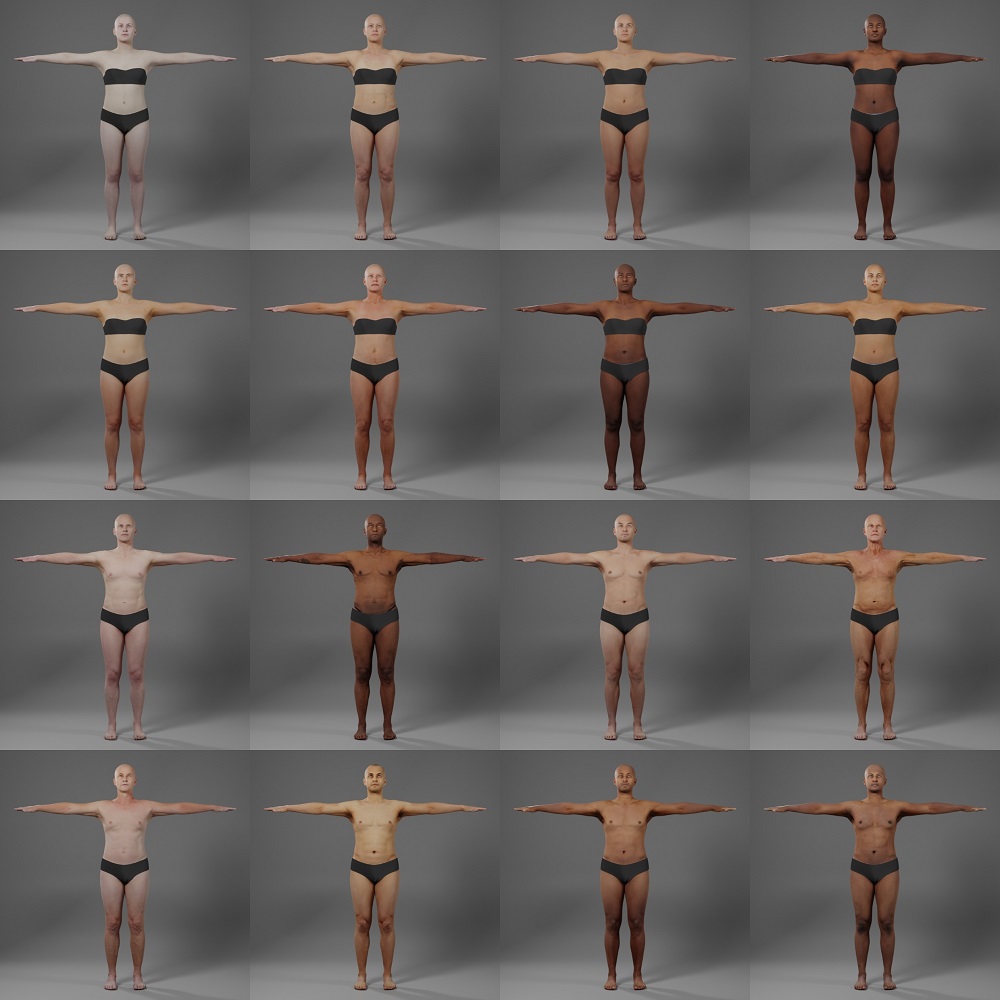}
        \caption{\centering Body.}
         \label{fig:body_skin_textures}
    \end{subfigure}
    \caption{Examples of skin textures from our library.}
    \label{fig:skin_textures}
\end{figure}

For the face we use the high quality skin texture library of \textcite{wood2021fake}, as shown in \autoref{fig:face_skin_textures}.
For the body we use a set of 25 high quality textures from 3D body scans~\cite{ten24}, as shown in \autoref{fig:body_skin_textures}.
For each scan we extract albedo, displacement and an approximated bump map for high-frequency details in the SOMA UV space described above.

\begin{figure}
    \centering
    \centering
        \begin{subfigure}{0.32\linewidth}
        \includegraphics[width=\linewidth]{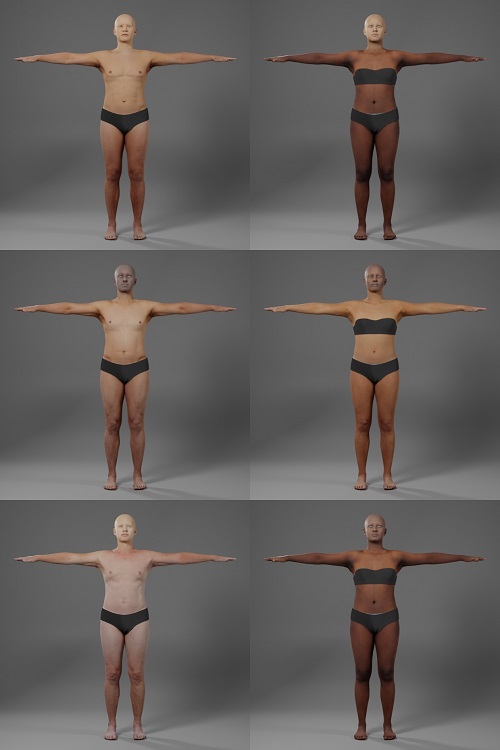}
        \caption{\centering Random sampling.}
        \label{fig:skin_matching:none}
    \end{subfigure}
    \begin{subfigure}{0.32\linewidth}
        \includegraphics[width=\linewidth]{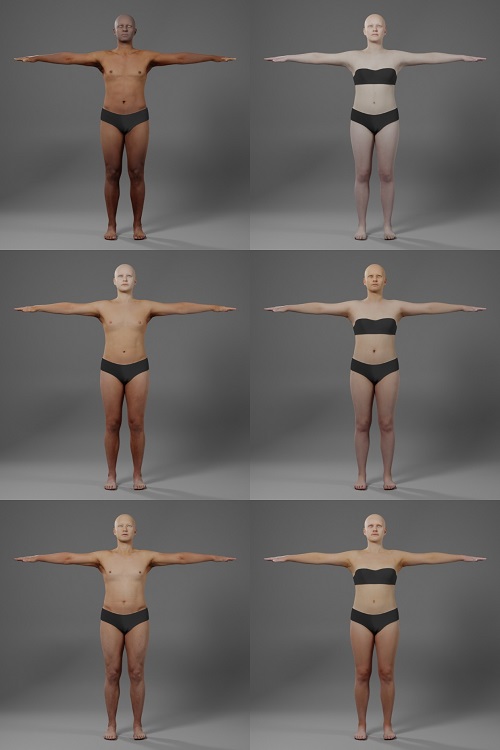}
        \caption{\centering Filtered sampling.}
        \label{fig:skin_matching:filter}
    \end{subfigure}
    \begin{subfigure}{0.32\linewidth}
        \includegraphics[width=\linewidth]{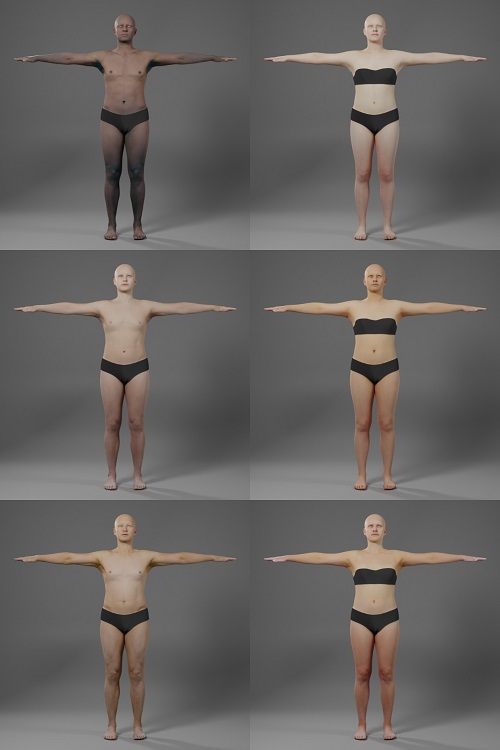}
        \caption{\centering Color adjustment.}
        \label{fig:skin_matching:match}
    \end{subfigure}
    \caption{Skin color matching process. We filter to achieve approximate matches when sampling then apply a further color correction.}
    \label{fig:skin_matching}
\end{figure}

Sampling face and body textures independently can result in significant mismatch in skin tone (see \autoref{fig:skin_matching:none}).

To address this we first sample a head texture from the library as our head texture library has greater diversity, then select a random body texture with average skin tone within some bound of perceptual similarity to that of the face using \autoref{eq:color} to determine perceptual color difference from input RGB values.
\begin{equation}
    \label{eq:color}
    \Delta C = \sqrt{\bigg(2 + \frac{\bar{r}}{256}\bigg) + \Delta R^2 + 4 \times \Delta G^2 + \bigg(2 + \frac{255 + \bar{r}}{256}\bigg) \times \Delta B^2}
\end{equation}
Where $\bar{r}$ is the average red component of the two colors. 

In general this provides quite close matches in skin tone between face and body, though there are still minor mismatches (see \autoref{fig:skin_matching:filter}). 
To address this we adjust the mean and variance of pixel values in the body texture to match that of the face texture, ensuring a quite precise match in skin tone of the body with the face (see \autoref{fig:skin_matching:match}).

\subsection{Clothing}

\textcite{wood2021fake} use mesh based assets for adding clothing and accessories to face.
For headwear, facewear (masks, eye-patches) and glasses the same technique can be used, simply parenting these assets to the head bone of the full body.
But other clothing items must now adapt to the dynamic pose of the body.
As such, we use displacement maps to model clothing items~\cite{ma2020learning}
We split these assets into tops and bottoms (including shoes), as well as using this technique to model some further accessories such as gloves, watches, bracelets and rings. 

Dynamic subdivision lets us produce very high fidelity results using this technique.
For each asset we author normal, roughness and metallic maps in addition to albedo and displacement, providing a high level of realism in terms of shading.

\begin{figure}
    \centering
    \includegraphics[width=\linewidth]{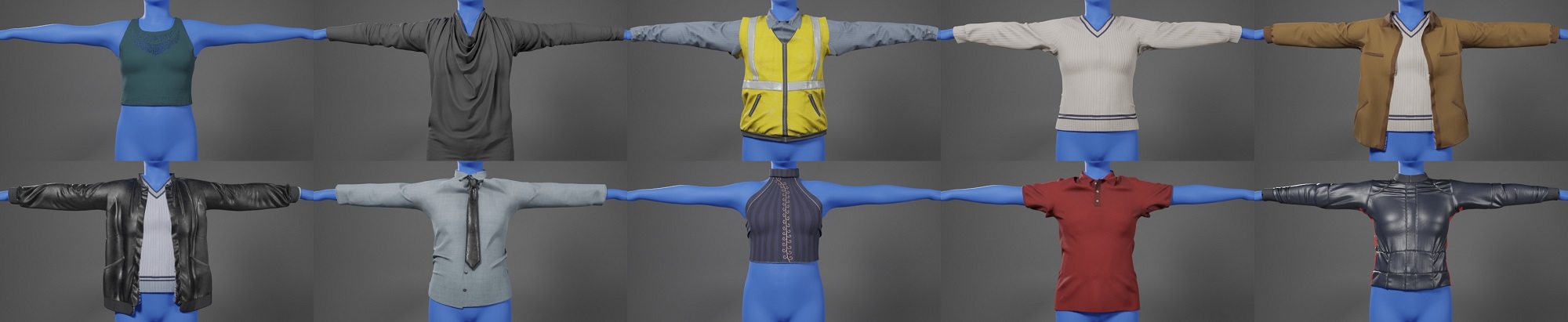}
    \includegraphics[width=\linewidth]{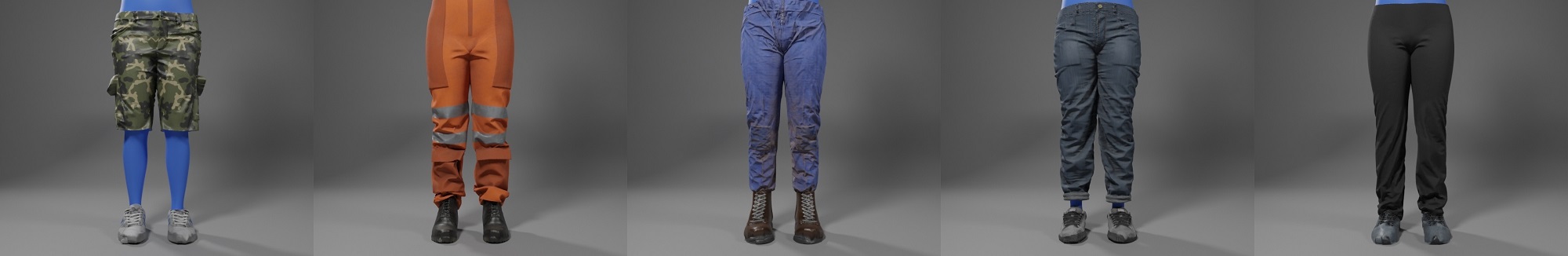}
    \includegraphics[width=\linewidth]{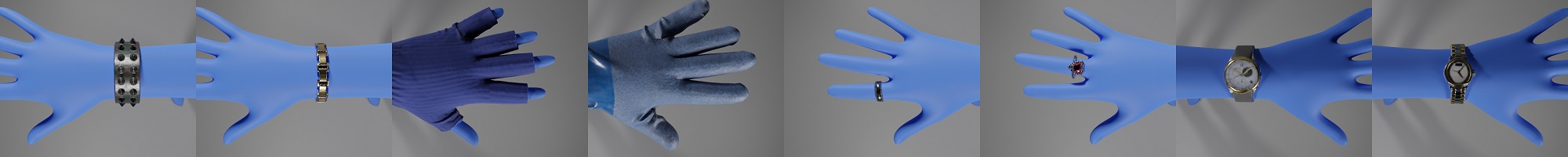}
    \caption{Examples of displacement clothing from our library including tops, bottoms, bracelets, gloves, ring and watches.}
    \label{fig:clothing_library}
\end{figure}

The clothing items are authored using Marvelous Designer~\cite{MarvelousDesigner} and displacement maps baked using Marmoset Toolbag~\cite{MarmosetToolbag}.
Manual cleanup and material detail is authored in Substance Painter~\cite{Substance}. Examples of some of the assets in our displacement map clothing library can be seen in \autoref{fig:clothing_library}.

There are a few significant shortcomings of this method, most obviously it is not possible to represent loose clothing using displacement maps.
Furthermore, simulation is impossible; the clothing must directly follow the body mesh underneath it when animated.
We find that displacement maps can give surprisingly compelling results for more than just very tight-fitting clothing as one might expect, but cannot be used for items such as dresses and skirts, and items like ties or jackets do not behave realistically in certain poses. 
To address these issues we plan to incorporate mesh based clothing in the future, along with cloth simulation.

\subsection{Pose Library}

For the face, the expression library of \textcite{wood2021fake} is used.
For the body we use the AMASS dataset~\cite{AMASS:ICCV:2019} as an initial pose library.
To this we add data collected using a motion-capture stage and processed using MoSh~\cite{loper2014mosh}.
Some of this motion-capture data is targeted specifically to fill gaps in the existing library such as poses with crossed legs.
In total our body pose library contains over 2 million frames at 30 fps, so approximately 19 hours of motion data.

In some cases the pose data is captured including articulated hand pose, but in many cases this is missing.
For frames without hand pose we randomly sample poses for each hand from the MANO dataset~\cite{MANO:SIGGRAPHASIA:2017} and splice these on.
Face expression (and eye pose) and body pose are sampled independently and spliced together.
In general we find this produces very plausible results, particularly for single-frame (i.e., non-sequential) data.

When collecting motion-capture data it is common to start (and end) the motion sequence in a canonical pose, often T-pose.
As a result we found that when sampling poses uniformly we had a very high occurrence of these T-poses, as such we use a Gaussian mixture model (GMM) to classify poses into a set of coarse classes one of which is T-pose.
This allows us to significantly down-weight T-poses in our resulting samples.

In addition, we found relatively neutral, standing poses were common and typically not useful when it comes to training DNNs for downstream tasks such as landmark detection.
Consequently, we also up-weight frames with higher mean absolute joint angles, i.e., frames which we consider to have more `interesting' poses.
We employ a similar approach for sampling facial expressions from the expression library of \textcite{wood2021fake}, weighted by mean blendshape activation.
Finally, we randomly mirror body poses to effectively double the number of unique poses.

\begin{figure}
    \centering
    \includegraphics[width=\linewidth]{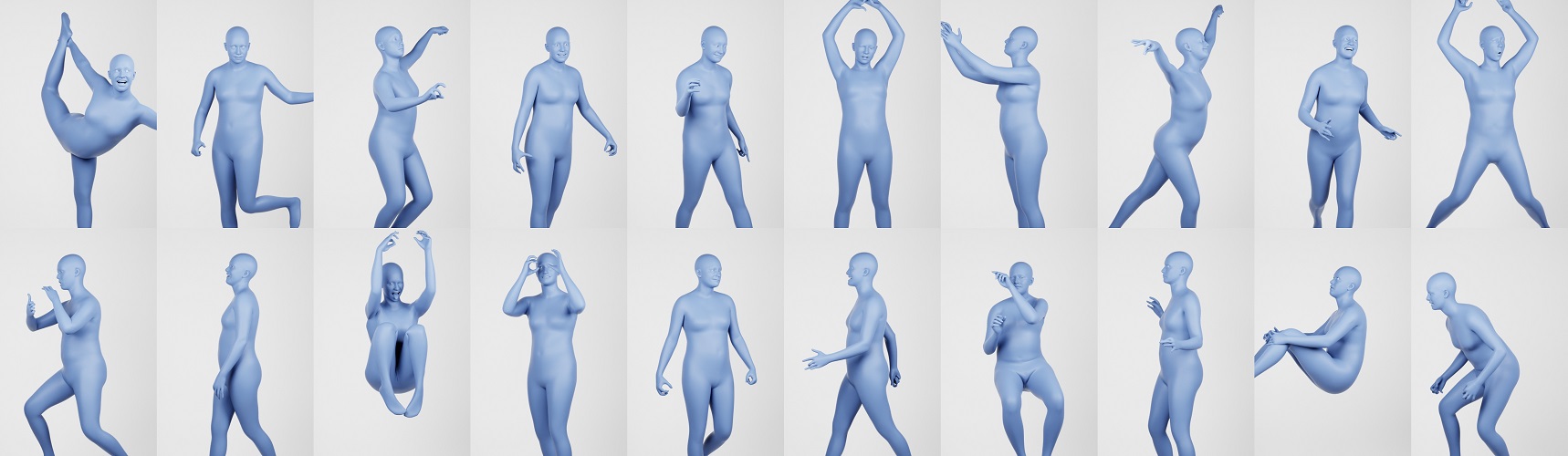}
    \caption{Example poses sampled from our body pose library with spliced facial expressions from \textcite{wood2021fake} and hand poses from MANO~\cite{MANO:SIGGRAPHASIA:2017} in some frames.}
    \label{fig:pose_library}
\end{figure}

Examples of poses sampled from our library using the above technique are shown in \autoref{fig:pose_library}.

%% file: landmarks.tex
\begin{figure}
    \centering
        \begin{subfigure}{0.48\linewidth}
        \includegraphics[width=\linewidth]{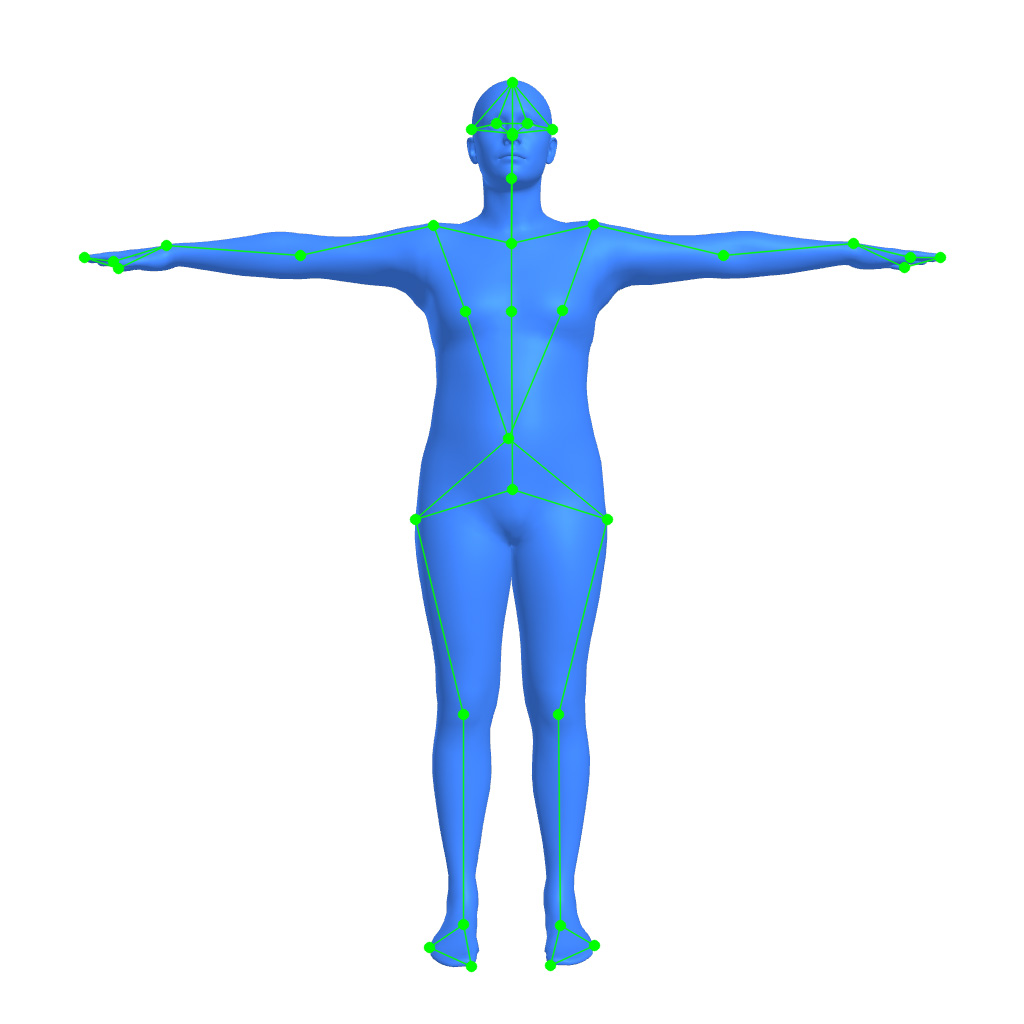}
        \caption{\centering Sparse (36).}
        \label{fig:landmarks_sparse}
    \end{subfigure}
    \begin{subfigure}{0.48\linewidth}
        \includegraphics[width=\linewidth]{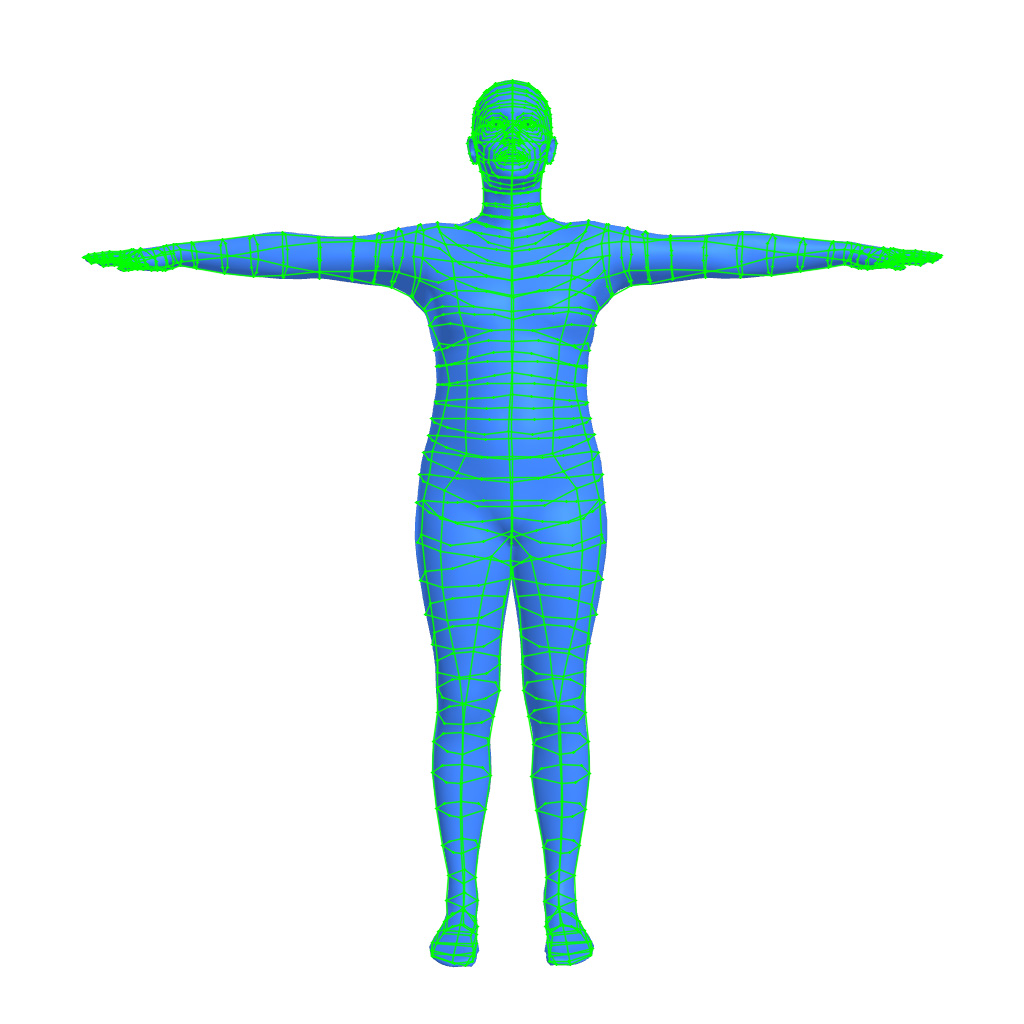}
        \caption{\centering Dense (1428).}
        \label{fig:landmarks_dense}
    \end{subfigure}
    \caption{Full-body landmark sets used for (a) detection and (b) dense tracking.}
    \label{fig:landmark_defs}
\end{figure}

Perhaps one of the most common use-cases for this kind of human-centred visual data is detection and tracking of people within images.
As such, we define landmark definitions corresponding to vertices of our body model defined in \autoref{sec:shape_model}.
A sparse definition of just 36 landmarks (\autoref{fig:landmarks_sparse}) which is used for detection and tracking with the sliding window approach outlined by \textcite{wood2022dense}.
As well as a dense definition of 1428 landmarks (\autoref{fig:landmarks_dense}) used for model fitting, see \autoref{sec:model_fitting}.
Using these definitions it is trivial to generate 2D landmark annotations for our synthetic data using the vertex location outputs (\autoref{fig:ground_truth:vertices}).

We render a dataset of 100,000 images containing a single person, with 20,000 identities and 5 frames per identity, using the pipeline outlined in \autoref{sec:render_pipeline}.
Each frame contains different pose and environmental lighting to increase the diversity of the data.
An example of such an input image used for training is shown in \autoref{fig:ground_truth:rgb} above.

To regress sparse landmarks we train a MobileNetV2~\cite{sandler2018mobilenetv2} model, for dense landmarks we train a ResNet101~\cite{he2016resnet} model, both with $256 \times 256$ pixel input image size.
In both cases we train using the procedure of \textcite{wood2022dense} with Gaussian Negative Log Likelihood (GNLL) loss and heavy use of data augmentation techniques.
The models therefore predict 2D landmark positions as well as per-landmark uncertainty values.

\subsection{Hand and face sub-networks}

When predicting landmarks for the full body as described above we find performance for the hands is poor.
This is not surprising given how small the hand is in the 256 pixel ROI.
The shape of the hand and how it moves also result in high levels of self-occlusion, making this task especially challenging.
Consequently, we train dedicated DNNs for hand landmark prediction using a ROI including just the hand as input.

\begin{figure}
    \centering
    \includegraphics[width=\linewidth]{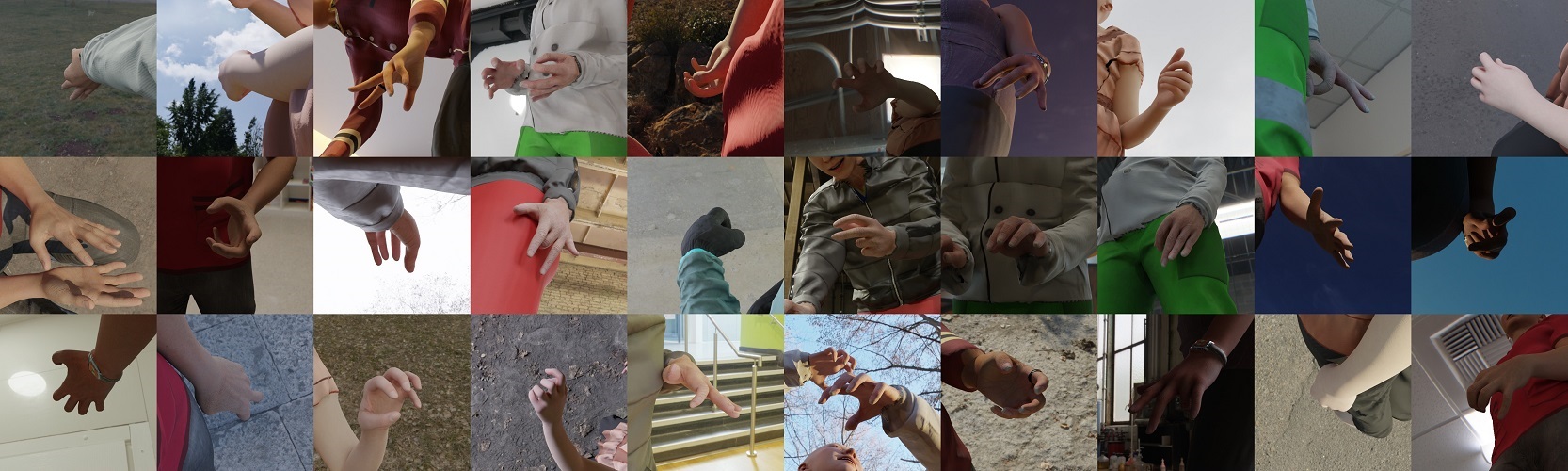}
    \caption{Sample images from our hand dataset.}
    \label{fig:hand_data}
\end{figure}

\begin{figure}
    \centering
        \begin{subfigure}{0.48\linewidth}
        \includegraphics[width=\linewidth]{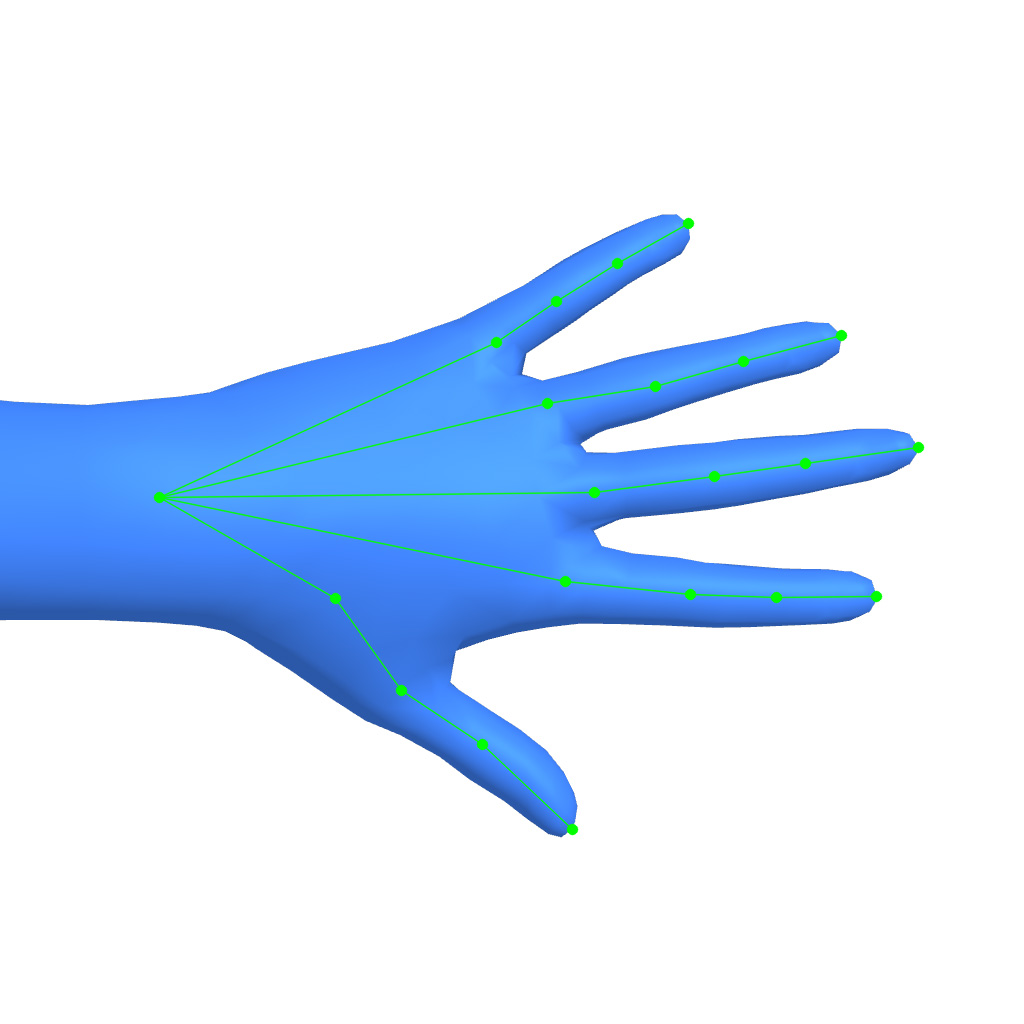}
        \caption{\centering Sparse hand (21).}
    \end{subfigure}
    \begin{subfigure}{0.48\linewidth}
        \includegraphics[width=\linewidth]{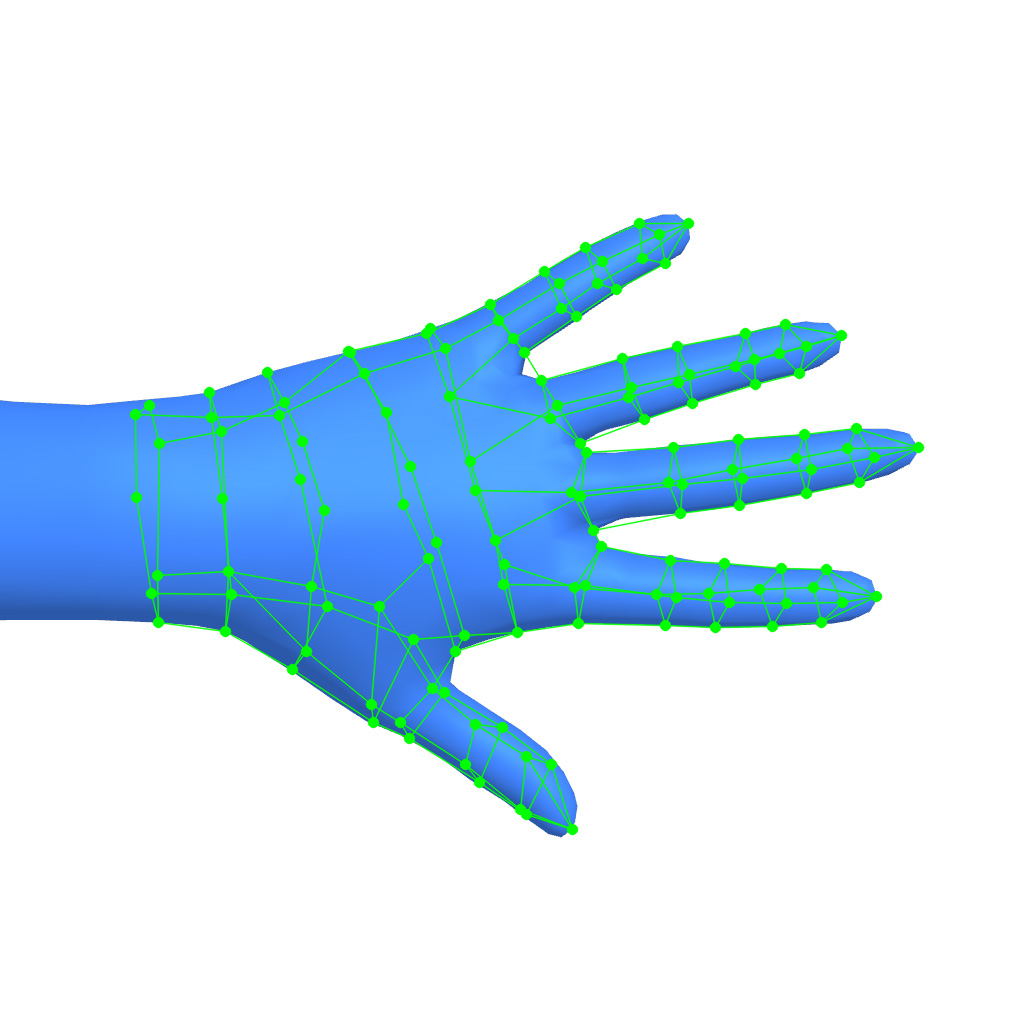}
        \caption{\centering Dense hand (141) .}
    \end{subfigure}
    \caption{Hand landmark sets used for (a) detection and (b) dense tracking.}
    \label{fig:hand_landmark_defs}
\end{figure}

We generate a dataset of 100,000 synthetic images cropped to include just the left hand, examples are shown in \autoref{fig:hand_data}.
We also define sparse and dense landmark definitions for just the hands shown in \autoref{fig:hand_landmark_defs}.
In the dense case the hand landmarks are a subset of the full-body definition in \autoref{fig:landmarks_dense} meaning they have direct correspondence.

Again we train using the procedure of \textcite{wood2022dense}, using MobileNetV2 \cite{sandler2018mobilenetv2} in the sparse case and ResNet18~\cite{he2016resnet} in the dense case with $128 \times 128$ pixel input image size.
We increase the amount of rotation augmentations used to further increase data diversity.
We also increase the frequency of motion blur augmentation to match observations in real data due to the typically faster motion of the hands than other body parts.

At run-time we first predict full-body landmarks and use these to extract and approximate ROI around the hand. 
We then use the sparse DNN to iteratively refine the ROI and finally run the dense DNN to get output landmarks.
Due to the direct correspondence in the dense definition we can overwrite the hand landmarks from the initial prediction, interpolating at the wrist. 

As our network has only seen left hands, and our initial full-body prediction disambiguates the left and right hands, we simply mirror the ROI for the right hand and input it to our left hand landmark DNNs.
The returned landmarks are then mirrored back before use.

Similar to hands, we find that face landmarks are also not predicted accurately when regressing full-body landmarks with a single DNN.
This is again likely to be due to the small size of the face in the 256 pixel ROI used in those models.
As such, we also take a dedicated face ROI and use the DNN of \textcite{wood2022dense} to regress accurate face landmarks.
As the face model used is the same, the face landmarks also retain a direct correspondence, so we can again overwrite those of the initial prediction.

\subsection{Results}

\begin{figure}
    \centering
    \includegraphics[width=\linewidth]{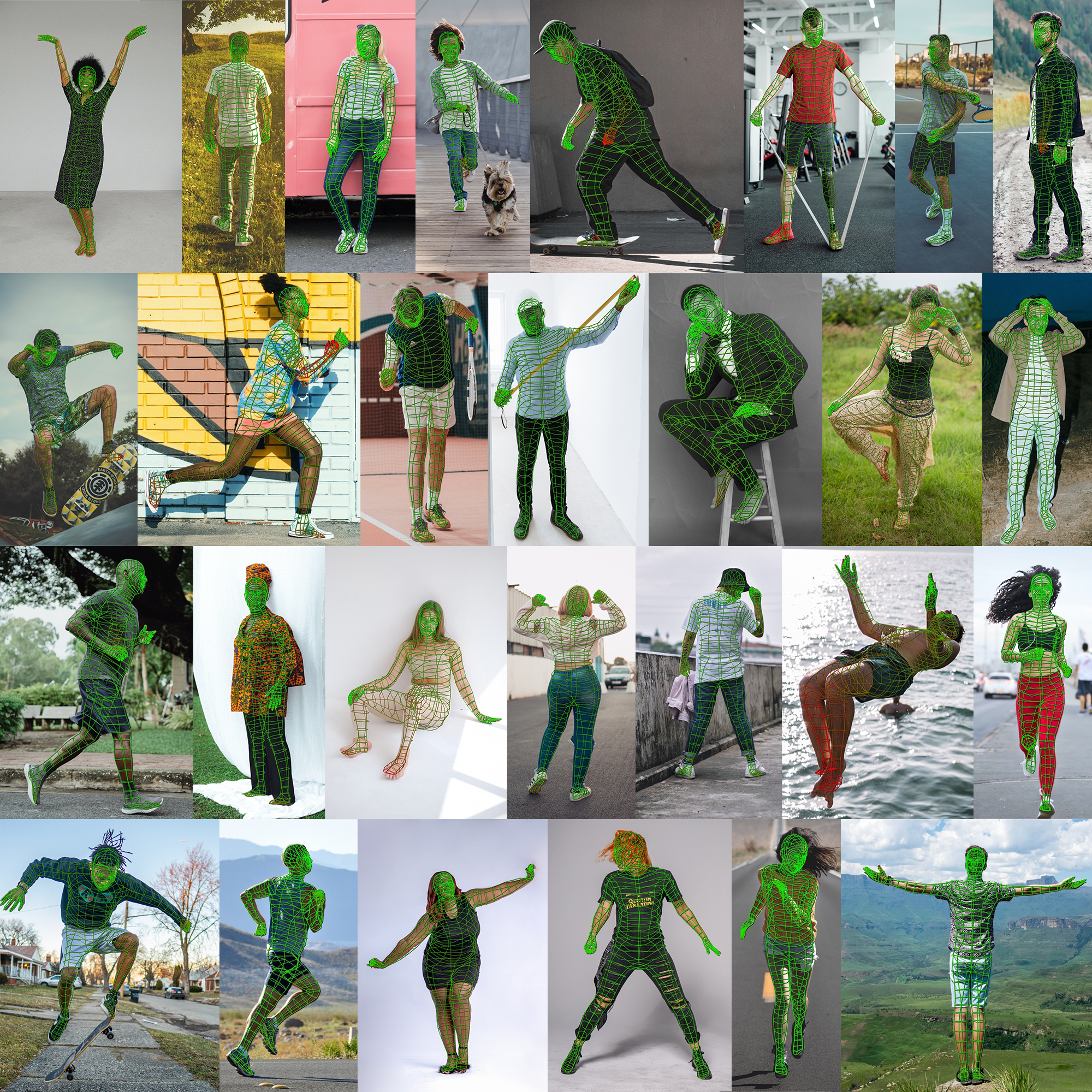}
    \caption{Dense landmark tracking results. Confidence is colour coded, with green being high and red being low. Images collected from \url{https://pexels.com}.}
    \label{fig:landmark_predictions}
\end{figure}

Some results of our dense full body landmark prediction method (including hand sub-networks) are shown in \autoref{fig:landmark_predictions}.
We are able to deal with a large range of pose, shape, appearance and environment.
In some cases we are even able to deal with loose clothing, children and prosthetic limbs despite these not being modelled in our synthetic data.
Particularly useful is the ability of the network to predict plausible landmarks with high uncertainty in cases of partial occlusion, like when one arm is totally hidden by the body.

\begin{figure}
    \centering
    \includegraphics[width=\linewidth]{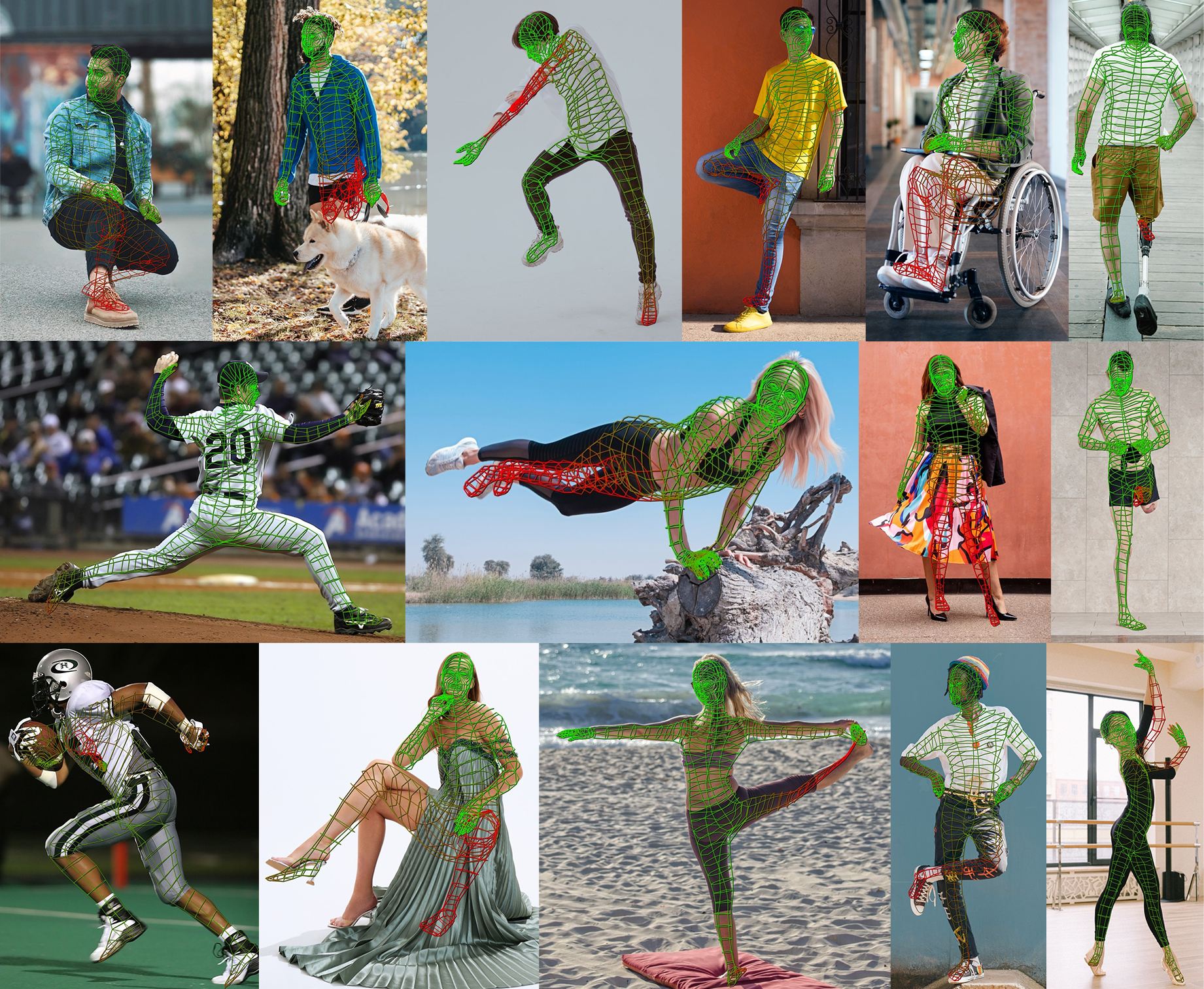}
    \caption{Examples of failures of our landmark prediction method. Confidence is colour coded, with green being high and red being low. Images collected from \url{https://pexels.com}.}
    \label{fig:landmark_failures}
\end{figure}

Examples of failures of our method are shown in \autoref{fig:landmark_failures}.
We particularly struggle with extreme poses, heavy (self-)occlusion, loose clothing, missing and prosthetic limbs.
This is likely because many of these elements are not currently modelled explicitly in our synthetics data generation pipeline.
In fact, missing limbs cannot even be represented by our parametric body model described in \autoref{sec:shape_model}.
However, due to our use of GNLL loss and resulting prediction of uncertainties, we typically get an accurate indication of where these errors are occurring, as demonstrated in many of these examples.
Future work on our shape model and rendering pipeline will aim to fill these gaps in representation of real world data.

%% file: model_fitting.tex
It is often useful not just to have landmarks, but to have a parameterized representation of a person's shape and motion .
We extend the approach of \textcite{wood2022dense} from face reconstruction to fit our complete body model described in \autoref{sec:shape_model} to the dense landmarks output by the pipeline of \autoref{sec:landmarks}.

So, given probabilistic dense 2D landmarks $L$, our goal is to find optimal model parameters $\myvec{\Phi}^*$ that minimize the following energy:

\begin{multline*}
E(\myvec{\Phi}; L) =
\underbrace{
    E_{
        \vphantom{identity}
        \textrm{landmarks}%
    }
}_{
    \textrm{Data term}
}
+ \\
\underbrace{
    E_{\textrm{face\_identity}} +
    E_{\textrm{body\_identity}} +
    E_{\textrm{expression}} +
    E_{\textrm{pose}} +
    E_{\textrm{temporal}} +
    E_{\textrm{intersect}}
}_{
    \textrm{Regularizers}
}
\end{multline*}

Where we are optimizing $\myvec{\Phi}$, that is face identity $\vec{\gamma}$, body identity $\vec{\beta}$, expression for each frame $\myvec{\Psi}$, pose for each frame $\myvec{\Theta}$, and camera rotations $\myvec{R}$ and positions $\myvec{T}$.

\begin{equation*}
\myvec{\Phi} = \{
\underbrace{
    \vec{\gamma},
    \vec{\beta},
    \myvec{\Psi}_{F \times |\vec{\psi}|},
    \myvec{\Theta}_{F \times|\vec{\theta}|}
}_{\textrm{Human\vphantom{(}}}
;\,
\underbrace{
    \vphantom{\myvec{\Psi}_{F \times |\vec{\psi}|}}
    \myvec{R}_{C \times 3},
    \myvec{T}_{C \times 3}
}_{\textrm{Cameras}}
\}
\end{equation*}

Where $F$ is the number of frames in the given sequence and $C$ is the number of cameras.
$E_{\textrm{landmarks}}$ takes the same form as in \textcite{wood2022dense}.
$E_{\textrm{face\_identity}}$, $E_{\textrm{expression}}$, $E_{\textrm{temporal}}$ and $E_{\textrm{intersect}}$ also follow the implementation of \textcite{wood2022dense}.

For $E_{\textrm{body\_identity}}$ we use an L2 prior given that SMPL-H uses a variance-scaled PCA basis for identity.
It may be beneficial to use a GMM body identity prior (as we do for the face) to promote more plausible body shape, we leave this is a potential direction for future work.

For pose, instead of using an L2 prior as in \textcite{wood2022dense}, we use three GMM priors.
One for body pose (excluding hand and eye pose) with a GMM fit to a subset of our pose library, and one for each hand with the GMMs fit to the MANO dataset~\cite{MANO:SIGGRAPHASIA:2017} for each hand respectively. 
This helps to promote plausible poses in a data-driven way.
More advanced pose priors (e.g., DNN) could provide better results~\cite{pavlakos2019expressive}, again we leave this for future work.

In cases of 2D-to-3D lifting such as this, bodies provide a much harder challenge than faces.
After projection there is much more ambiguity in limb position, for example, due to the high range of motion of some body parts compared to the face.
Further, self-occlusion is much more common for bodies, while for faces symmetry provides a very strong prior when limited self-occlusion does occur.
As such, we observe that effective model fitting is much more difficult than for faces, and the monocular case is often ill-posed.
We find that multiple camera views ($C \ge 3$) are required and results improve significantly for higher numbers of cameras.

We also find reasonable initialization to be more important than for faces, and significantly harder.
For faces simple 6-DoF alignment using PnP is sufficient to get a very good starting point for the optimizer, but in the case of the full body this gives quite poor results.
Particularly for fine details like hand pose which will struggle to converge without good initialization, even if the landmarks are highly accurate.

To address this we initialize the pose using a machine learning approach to predict pose directly from one of the available views~\cite{ExPose:2020}, providing the optimizer with a very good starting point.
The multi-view landmarks are then used by the optimizer to achieve highly accurate 3D consistency across views, and so a very precise registration in world-space.
Many machine learning approaches do not take multi-view data as input and, even when they do, struggle to achieve very precise alignment and consistency between views as required here.
Future work might attempt to improve this initialization step to use multi-view data and reduce the need for the secondary optimization step.

\begin{figure}
    \centering
    \includegraphics[width=\linewidth]{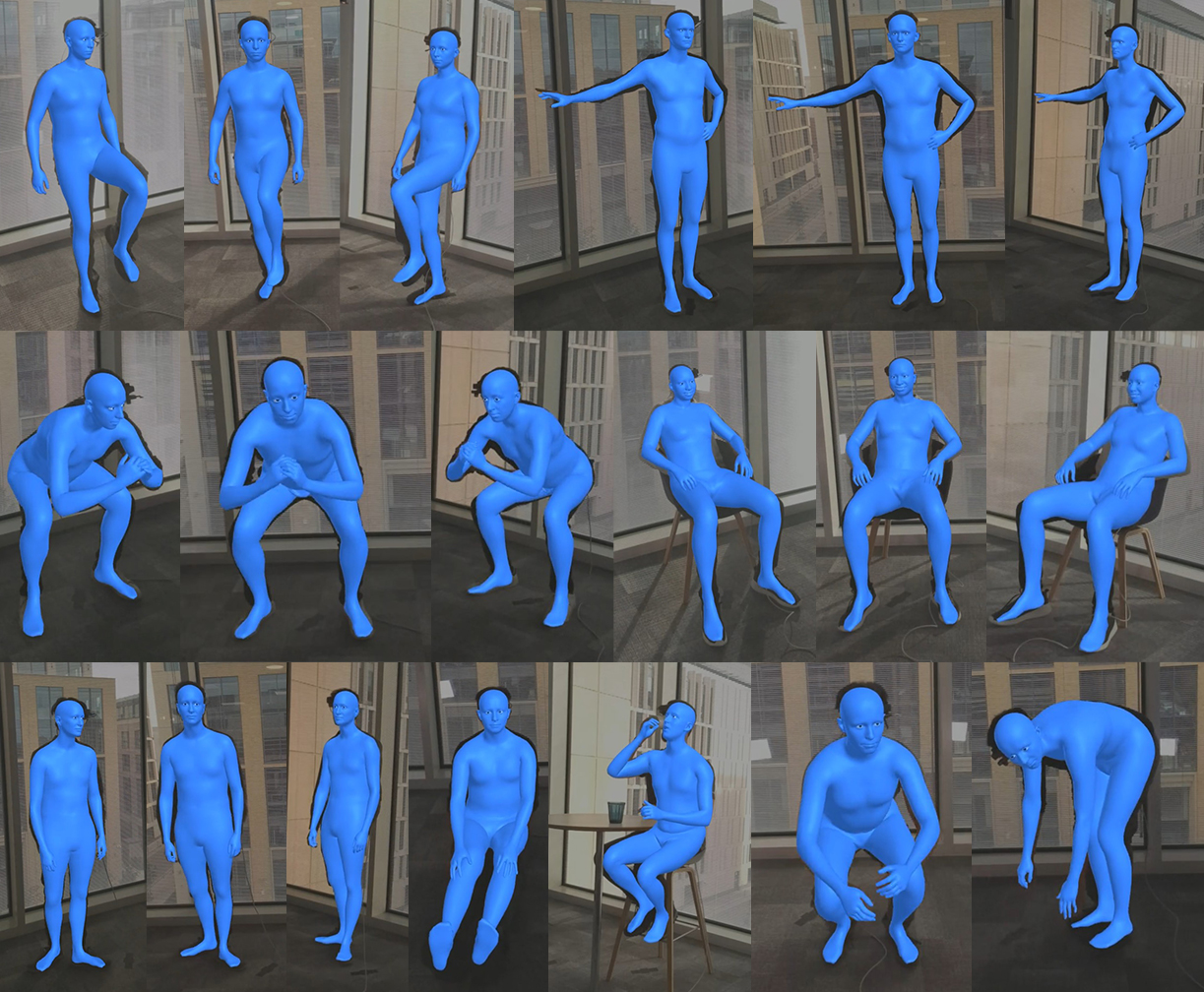}
    \caption{Example model fitting results on data collected using three Azure Kinect RGB cameras. Showing three viewpoints for various subjects and poses, and a single viewpoint for four diverse poses in the bottom right.}
    \label{fig:fitting_results}
\end{figure}

Some results of our model fitting approach are shown in \autoref{fig:fitting_results}.

%% file: references.bib
@misc{wood2022dense,
    doi = {10.48550/ARXIV.2204.02776},
    url = {https://arxiv.org/abs/2204.02776},
    author = {Wood, Erroll and Baltrusaitis, Tadas and Hewitt, Charlie and Johnson, Matthew and Shen, Jingjing and Milosavljevic, Nikola and Wilde, Daniel and Garbin, Stephan and Raman, Chirag and Shotton, Jamie and Sharp, Toby and Stojiljkovic, Ivan and Cashman, Tom and Valentin, Julien},
    title = {3D face reconstruction with dense landmarks},
    publisher = {arXiv},
    year = {2022},
    copyright = {Creative Commons Attribution Non Commercial Share Alike 4.0 International}
}

@misc{wood2021fake,
    title={Fake It Till You Make It: Face analysis in the wild using synthetic data alone},
    author={Erroll Wood and Tadas Baltru\v{s}aitis and Charlie Hewitt and Sebastian Dziadzio and Matthew Johnson and Virginia Estellers and Thomas J. Cashman and Jamie Shotton},
    year={2021},
    eprint={2109.15102},
    archivePrefix={arXiv},
    primaryClass={cs.CV}
}

@article{loper2014mosh,
  title={MoSh: Motion and shape capture from sparse markers},
  author={Loper, Matthew and Mahmood, Naureen and Black, Michael J},
  journal={ACM ToG},
  volume={33},
  number={6},
  pages={1--13},
  year={2014},
  publisher={ACM New York, NY, USA}
}

@inproceedings{ExPose:2020,
    title= {Monocular Expressive Body Regression through Body-Driven Attention},
    author= {Choutas, Vasileios and Pavlakos, Georgios and Bolkart, Timo and Tzionas, Dimitrios and Black, Michael J.},
    booktitle = {ECCV},
    year = {2020},
    url = {https://expose.is.tue.mpg.de}
}

@article{MANO:SIGGRAPHASIA:2017,
    title = {Embodied Hands: Modeling and Capturing Hands and Bodies Together},
    author = {Romero, Javier and Tzionas, Dimitrios and Black, Michael J.},
    journal = {ACM ToG},
    volume = {36},
    number = {6},
    series = {245:1--245:17},
    month = nov,
    year = {2017},
    month_numeric = {11}
}

@article{SMPL:2015,
    author = {Loper, Matthew and Mahmood, Naureen and Romero, Javier and Pons-Moll, Gerard and Black, Michael J.},
    title = {{SMPL}: A Skinned Multi-Person Linear Model},
    journal = {ACM ToG},
    month = oct,
    number = {6},
    pages = {248:1--248:16},
    publisher = {ACM},
    volume = {34},
    year = {2015}
}

@Misc{CyclesRenderer,
    author = {{Blender Foundation}},
    title = {Cycles Renderer},
    howpublished = {\url{https://www.cycles-renderer.org/}},
    year = {2021}
}

@misc{MarvelousDesigner,
    author={{CLO} Virtual Fashion Inc.},
    title={Marvelous Designer},
    howpublished={\url{https://marvelousdesigner.com/}},
    year={2022}
}

@misc{Substance,
    author={Adobe Inc.},
    title={Substance Painter},
    howpublished={\url{https://www.adobe.com/products/substance3d-painter.html}},
    year={2022}
}

@misc{MarmosetToolbag,
    author={Marmoset {LLC}},
    title={Marmoset Toolbag},
    howpublished={\url{https://marmoset.co/Toolbag/}},
    year={2022}
}

@conference{AMASS:ICCV:2019,
  title = {{AMASS}: Archive of Motion Capture as Surface Shapes},
  author = {Mahmood, Naureen and Ghorbani, Nima and Troje, Nikolaus F. and Pons-Moll, Gerard and Black, Michael J.},
  booktitle = {ICCV},
  pages = {5442--5451},
  month = oct,
  year = {2019},
  month_numeric = {10}
}

@inproceedings{sandler2018mobilenetv2,
  title={Mobilenetv2: Inverted residuals and linear bottlenecks},
  author={Sandler, Mark and Howard, Andrew and Zhu, Menglong and Zhmoginov, Andrey and Chen, Liang-Chieh},
  booktitle={CVPR},
  pages={4510--4520},
  year={2018}
}

@INPROCEEDINGS{he2016resnet,
  author={K. {He} and X. {Zhang} and S. {Ren} and J. {Sun}},
  booktitle={CVPR}, 
  title={{Deep Residual Learning for Image Recognition}}, 
  year={2016},
}

@inproceedings{bae2023digiface1m,
  title={DigiFace-1M: 1 Million Digital Face Images for Face Recognition},
  author={Bae, Gwangbin and de La Gorce, Martin and Baltru{\v{s}}aitis, Tadas and Hewitt, Charlie and Chen, Dong and Valentin, Julien and Cipolla, Roberto and Shen, Jingjing},
  booktitle={WACV},
  year={2023},
  organization={IEEE}
}

@inproceedings{sun2022shift,
  title={SHIFT: A Synthetic Driving Dataset for Continuous Multi-Task Domain Adaptation},
  author={Sun, Tao and Segu, Mattia and Postels, Janis and Wang, Yuxuan and Van Gool, Luc and Schiele, Bernt and Tombari, Federico and Yu, Fisher},
  booktitle={CVPR},
  pages={21371--21382},
  year={2022}
}

@inproceedings{pavlakos2019expressive,
  title={Expressive body capture: 3d hands, face, and body from a single image},
  author={Pavlakos, Georgios and Choutas, Vasileios and Ghorbani, Nima and Bolkart, Timo and Osman, Ahmed AA and Tzionas, Dimitrios and Black, Michael J},
  booktitle={CVPR},
  pages={10975--10985},
  year={2019}
}

@misc{thomas_pandikow_kim_stanley_grieve_2021,
    title={Open Synthetic Dataset for Improving Cyclist Detection},
    url={https://paralleldomain.com/},
    journal={https://paralleldomain.com/open-datasets/bicycle-detection},
    publisher={Parallel Domain},
    author={Thomas, Phillip and Pandikow, Lars and Kim, Alex and Stanley, Michael and Grieve, James},
    year={2021},
    month={11}
}

@inproceedings{mcduff2021synthetic,
  title={Synthetic Data for Multi-Parameter Camera-Based Physiological Sensing},
  author={McDuff, Daniel and Liu, Xin and Hernandez, Javier and Wood, Erroll and Baltrusaitis, Tadas},
  booktitle={EMBC},
  pages={3742--3748},
  year={2021},
  organization={IEEE}
}

@misc{ten24,
    author={Ten24 Animation Ready Body Scans.},
    title={Ten24 Media LTD},
    howpublished={\url{https://www.3dscanstore.com/retopologised-body-models/animation-ready-body-scans}},
    year={2022}
}

@inproceedings{ma2020learning,
  title={Learning to dress 3d people in generative clothing},
  author={Ma, Qianli and Yang, Jinlong and Ranjan, Anurag and Pujades, Sergi and Pons-Moll, Gerard and Tang, Siyu and Black, Michael J},
  booktitle={CVPR},
  pages={6469--6478},
  year={2020}
}
